\pdfoutput=1

\documentclass[11pt]{article}

\usepackage[]{acl}

\usepackage{times}
\usepackage{latexsym}
\usepackage{amsfonts}
\usepackage[T1]{fontenc}

\usepackage[utf8]{inputenc}

\usepackage{microtype}

\usepackage{inconsolata}

\usepackage{graphicx}
\usepackage{array}
\usepackage{amsmath}
\usepackage{caption}
\usepackage{subcaption}
\usepackage{booktabs}
\usepackage{multirow}
\title{Exploring Intrinsic Language-specific Subspaces in Fine-tuning Multilingual Neural Machine Translation}

\author{
    Zhe Cao, Zhi Qu, Hidetaka Kamigaito, Taro Watanabe \\
    Nara Institute of Science and Technology \\
    \texttt{\{cao.zhe.bw4, zhi.qu.pv5, kamigaito.h, taro\}@is.naist.jp} 
}

\begin{document}
\maketitle
\begin{abstract}
Multilingual neural machine translation models support fine-tuning hundreds of languages simultaneously. However, fine-tuning on full parameters solely is inefficient potentially leading to negative interactions among languages. In this work, we demonstrate that the fine-tuning for a language occurs in its intrinsic language-specific subspace with a tiny fraction of entire parameters. Thus, we propose language-specific LoRA to isolate intrinsic language-specific subspaces. Furthermore, we propose architecture learning techniques and introduce a gradual pruning schedule during fine-tuning to exhaustively explore the optimal setting and the minimal intrinsic subspaces for each language, resulting in a lightweight yet effective fine-tuning procedure. The experimental results on a 12-language subset and a 30-language subset of FLORES-101 show that our methods not only outperform full-parameter fine-tuning up to 2.25 spBLEU scores but also reduce trainable parameters to $0.4\%$ for high and medium-resource languages and $1.6\%$ for low-resource ones. Code will be released at \url{https://github.com/Spike0924/LSLo}.

\end{abstract}

\section{Introduction}
Multilingual Neural Machine Translation (MNMT) aims to use a single model to translate among different languages \cite{ha-etal-2016-toward,10.1162/tacl_a_00065}. Recent studies of MNMT \citep{fan2020englishcentric,nllbteam2022language} achieved significant progress in training large-scale pre-trained models supporting hundreds of languages. Benefiting from cross-language learning, these pre-trained models offer the possibility of fine-tuning with limited data and show better performance in low-resource languages and non-English directions. However, multilingual fine-tuning still suffers from two limitations: (1) full-parameter fine-tuning becomes inefficient as the model size increases; (2) negative interactions among languages \citep{duh-etal-2012-learning,mohammadshahi-etal-2022-small,he-etal-2023-gradient,chen2023pareto,huang-etal-2023-towards} lower the performance of high-resource languages. 


Recent studies have shown that the fine-tuning of pre-trained models can be re-parameterized in an intrinsic subspace, i.e., a low-rank subspace with tiny parameters \citep{li2018measuring,qin2022exploring,zhang-etal-2023-fine}. This insight implies that language-specific fine-tuning in pre-trained MNMT models happens within intrinsic language-specific subspaces, thus overcoming the aforementioned limitations: (1) intrinsic subspaces significantly reduce the required trainable parameters; (2) isolating the intrinsic subspaces among languages alleviates the negative interference in the multilingual representations. Therefore, in this work, we propose Language-Specific LoRA (LSLo), consisting of multiple LoRA \citep{hu2021lora} modules with sparse language-specific activation, to model such intrinsic subspaces.


Moreover, prior works \citep{qu2022adapting,pfeiffer-etal-2022-lifting,pires-etal-2023-learning} allocate the same number of parameters to different languages, which can yield sub-optimal setup because pre-trained models have already learned a substantial amount of knowledge from high-resource languages given the imbalance distribution of training data. We hypothesize that fine-tuning of high-resource languages can be done in a smaller subspace compared to low-resource languages. To exhaustively explore the minimal intrinsic subspaces for each language, we first reduce the rank for high-resource languages and then introduce unstructured pruning with a Gradual Pruning Schedule \citep{he-etal-2023-gradient} during fine-tuning.


However, determining the optimal structure of LSLo remains challenging. First, there are 2 cases when selecting the language-specific sub-module of each LSLo: selected by source language (source-indexed) and selected by target language (target-indexed). Furthermore, although we intuitively expect that high-resource languages require smaller subspaces, it's still insufficient for the complex multilingual setting.  These lead to the exponential increase in the possible architectures with the increase of the number of model layers and supported languages. Therefore, in this work, we use two architecture Learning techniques to avoid the tedious manual trial-and-error. We applied Weight Learning \citep{JMLR:v20:18-598,pires-etal-2023-learning} to determine whether each LSLo module should be source-indexed or target-indexed, given its interpretability and ease of visualization. We also propose a Layer-wise Cross-Language Pruning method, which combines the LoRA modules of all languages at every layer for pruning to estimate the required subspace size for each language.

We conduct our experiments on a 12-language subset of FLORES-101 \citep{goyal2021flores101}. Results show that in a pre-trained MNMT model, the size of intrinsic language-specific subspace is highly correlated with the language's resource type. Specifically, High-resource languages can be fine-tuned within a very small parameter subspace. Our fine-tuning method outperforms full parameter fine-tuning by 1.3 spBLEU while only using $0.4\%$ trainable parameters for high and medium languages, and $1.6\%$ for low-resource ones. We further evaluate our method on a 30-language subset, achieving a 2.25 spBLEU improvement over full parameter fine-tuning with only 7\% trainable parameters, which demonstrates the efficiency and effectiveness of our method.
\section{Background}
Given a set of $\mathit{n}$ languages $\mathbb{L} = \{l_1,l_2,\cdots,l_n\}$, the multilingual translation task is defined as translating an input in source language $\mathit{src} \in \mathbb{L}$ into an output in target language $\mathit{tgt} \in \mathbb{L}$. To train an MNMT model, we need a parallel corpus including translations aligned at the sentence level for creating MNMT datasets. 
For instance, consider a collection with $m$ sets of sentences $\mathbb{S} = \{\mathbb{S}_1,\mathbb{S}_2, \cdots, \mathbb{S}_m\}$, each sentence set includes sentences in different languages sharing the same semantics, $\mathbb{S}_k=\{\boldsymbol{s}_{l_1}^k,\boldsymbol{s}_{l_2}^k,\cdots, \boldsymbol{s}_{l_n}^k\}$. With a parallel corpus, we can conveniently construct MNMT datasets including different translation directions $\mathit{src} \rightarrow \mathit{tgt}$ by choosing source and target sentences pairs from $\mathbb{S}$, e.g., $\boldsymbol{s}_\mathit{src}^k$ as the input $\boldsymbol{x}$ and $\boldsymbol{s}_\mathit{tgt}^k$ as the output $\boldsymbol{y}$ of a single translation pair $(\boldsymbol{x}, \boldsymbol{y})$. Given a MNMT dataset with $N$ translation pairs $\mathbb{D}=\{(\boldsymbol{x}_i,\boldsymbol{y}_i),i\in 1\cdots N\}$, the training loss is defined as:
\begin{equation}
    \label{loss}
    \mathcal{L}_{\mathit{MNMT}}=-\sum_{\boldsymbol{x},\boldsymbol{y} \in \mathbb{D}}\sum_{j=1}^{J} \log\ p_{\theta}(y_j|\boldsymbol{y}_{<j},\boldsymbol{x})
\end{equation}
where $\boldsymbol{x}=x_1,x_2,\cdots,x_I$ is a source sentence with length $I$ and $\boldsymbol{y}=y_1,y_2,\cdots,y_J$ is the corresponding target sentence with length $J$. We say an MNMT model is English-centric if all language pairs in its training data include English. Models without this limitation are classified as many-to-many models.  In this work, we conduct experiments in a many-to-many setting.
\section{Methodology}
\subsection{Language-specific LoRA}
LoRA \citep{hu2021lora} is widely used in Parameter-efficient Fine-tuning (PEFT) for Large Language Models where fine-tuning is re-parameterized in a low-rank intrinsic subspace. For a weight matrix in a pre-trained model $\mathbf{W} \in \mathbb{R}^{d\times k}$, LoRA forward pass can be calculated as:
\begin{equation}
    \label{lora_def}
    \boldsymbol{h} = \mathbf{W}\boldsymbol{x} + \mathbf{BA}\boldsymbol{x}
\end{equation}
where $\mathbf{B} \in \mathbb{R}^{d \times r}$ and $\mathbf{A} \in \mathbb{R}^{r \times d}$. During training, $W$ will be frozen and the trainable parameters, i.e., $A$ and $B$, will be reduced from $d \times k$ to $d \times r + r \times k$, where $r \ll \mathit{min}(d, k)$.

In this work, we propose Language Specific LoRA (LSLo), an instance of Mixture-of-LoRAs \citep{feng-etal-2024-mixture-loras} but with a hard language-specific routing. Specifically, Each LSLo module contains multiple sparsely activated LoRA modules with different rank $r_{l_i}$ for each language, instead of sharing a unified parameter space across all languages. 
The forward pass of LSLo is calculated as:
\begin{equation}
    \label{lslo_def}
    \begin{aligned}
        \boldsymbol{h} &=\mathbf{W}\boldsymbol{x} + \mathrm{LSLo}(\boldsymbol{x},l_i) \\
        &= \mathbf{W}\boldsymbol{x} + \mathbf{B}_{l_i}\mathbf{A}_{l_i}\boldsymbol{x}
    \end{aligned}
\end{equation}
where $\mathit{l}_i \in \mathbb{L}$ is the selected language for this LSLo. Only the LoRA module of the selected language will be activated each time. Similar to LoRA, LSLo can be added to any weight matrices, e.g., projections for attention and fully connected layers. The number of trainable parameters for each LSLo module is $\sum_{i=1}^{n} (d \times r_{l_i} + r_{l_i} \times k)$, where $r_{l_i}$ is the reduced dimension for each language $l_i$. We are allowed to flexibly adjust the size of intrinsic language-specific subspaces through $r_{l_i}$, thus achieving higher parameter efficiency.
\subsection{Unstructured Pruning} \label{sec:GPS}
The assumption that higher-resource languages have smaller intrinsic subspaces naturally leads to the following question: How small can these subspaces be? Therefore, we adopt unstructured pruning \footnote{We directly use the implementation from PyTorch. \url{https://pytorch.org/docs/stable/generated/torch.nn.utils.prune.l1_unstructured.html}} to explore the minimal intrinsic language-specific subspaces exhaustively. Compared with structured pruning which directly removes entire rows or columns from a weight matrix, we choose unstructured pruning without the above limitation to achieve a higher pruning ratio. To ensure the stability of the model during training under a high pruning ratio, we introduce a Gradual Pruning Schedule \citep{zhu2017prune,he-etal-2023-gradient} during training. Specifically, the pruning ratio for each epoch $P_e$ is gradually increased from $0$ to the predefined target pruning ratio $P$. The entire training process is divided into three stages as denoted in equation \ref{gps}.
Given a predefined pruning ratio $P$ and the total training process has $T$ epochs. $E$ is the starting epoch for pruning, and the pruning process will last for $k$ epochs.
\begin{equation}
    \label{gps}
    P_e=
    \left\{
        \begin{array}{lc}
            0 & e \leq E \\
            P - P{(1-\frac{e-E}{k})}^3 & E < e \leq (E+k) \\
            P & (E+k) < e \leq T
        \end{array}
    \right.
\end{equation}
During the first $E$ epochs ($e \leq E$), no pruning is applied denoted by $P = 0$; for stage 2, the pruning ratio of the current $e$ epoch is gradually increased until reaching the target ratio $P$ for the next $k$ epochs; for stage 3, the pruning ratio $P$ is kept to the end.
For the following experiments, we empirically choose the start epoch $E=2$ and the duration $k=8$.
We provide a detailed description of the target pruning ratio $P$ used in each experiment, as well as the impact of different choices of $E$ and $k$ on performance in Appendix \ref{app:GPS}.
\section{Architecture Learning}
LSLo introduces additional hyperparameters: (1) each LSLo module can be selected by either the source or the target language; (2) each language can have a different rank $r_{l_i}$, leading to an exponential increase in possible architectures with the number of layers and languages. Therefore, we propose two architecture learning techniques in this section to avoid manual selection.
\subsection{Weight Learning}
\label{sec:WL}
Consider a translation from a source language $l_i \in \mathbb{L}$ to a target language $l_j \in \mathbb{L}$. We say an LSLo module is source-indexed if activated by the source language $l_i$ and is target-indexed if activated by the target language $l_j$.
Intuitively, we expect that the language information is transformed from the source side to the target side near the top layers of the encoder and the bottom layers of the decoder \citep{kudugunta-etal-2019-investigating}, which motivates the assumption that a layer in an encoder or decoder might prefer either source or target language, e.g., top layers of the encoder require target-indexed LSLo for more target side information while bottom layers of the encoder require source-indexed LSLo for more source side information. 
However, finding an optimal setting remains tedious work.
Inspired by Neural Architecture Search \citep{JMLR:v20:18-598,pires-etal-2023-learning}, we introduce a weight learning method here to determine the activation strategy for each layer's LSLo modules. We use $\mathit{mo}$ to denote any module where LSLo might be added, including the query, key, and value matrices of attention and cross-attention, as well as the up and down matrices of fully-connected layers. Given an LSLo module added to a pre-trained weight matrix $W$, let the layer index $W$ located is $i$, and the module $W$ belongs to is $\mathit{mo}$, we calculate weighted sum during forward pass as follows:
\begin{equation}
    \label{weight-learning}
    \begin{aligned}
        \boldsymbol{h}^i_{mo} = &\mathbf{W}^i_{mo}\boldsymbol{x}\ \ + \\
            &w^i_{\mathit{src}} \cdot \mathrm{LSLo}^i_{mo}(\boldsymbol{x}, l_{\mathit{src}})\ \ + \\
            &w^i_{\mathit{tgt}} \cdot \mathrm{LSLo}^i_{mo}(\boldsymbol{x}, l_{\mathit{tgt}}) \ \ .
    \end{aligned}
\end{equation}
where $w^i_{\mathit{src}}$, $w^i_{\mathit{tgt}}$ are shared among all LSLo modules in the same layer, and we use $\mathrm{softmax}$ to make sure the weights are non-negative and sum up to 1. We will simply choose the index strategy with the one having a larger weight.
\subsection{Intrinsic Subspace Estimation}\label{sec:intrinsic_est}

Intuitively, high-resource languages can be fine-tuned in smaller subspaces owing to the extensive knowledge learned during pre-training, while low-resource ones should preserve larger subspaces due to the limited resources. However, in practice, some medium-resource languages, such as Dutch, have data scales similar to high-resource languages, thus it is possible to reduce the size of subspaces. Additionally, some low-resource languages would benefit more from cross-lingual transfer thanks to their similarity to high-resource languages, e.g., the same language family, effectively allowing the reduction in the fine-tuning subspaces. Therefore, we propose an intrinsic subspace estimation technique using layer-wise cross-language pruning\footnote{We also use the implementation from PyTorch. \url{https://pytorch.org/docs/stable/generated/torch.nn.utils.prune.global_unstructured.html}} to comprehensively analyze the fine-tuning space demands for each language.

We apply LSLo to all possible weight matrices and group $\mathbf{B}$ matrices from LSLo modules of all languages in the same layer for pruning. We use the same unstructured pruning in Section \ref{sec:GPS}. Let $\#_\mathbf{B}$ be the number of parameters in matrix $\mathbf{B}$, $P_{\mathrm{ISE}}$ is the predefined pruning ratio, and $\#_{\mathrm{pruned}_\mathbf{B}}$ represents the actual number of parameters pruned from matrix $\mathbf{B}$. We measure the intrinsic subspace demands using the following importance score:
\begin{equation}
    \label{eq:IntrinsicEstimation}
    \mathrm{Score}(\mathbf{B}) = \#_{\mathrm{pruned}_\mathbf{B}} - P_{\mathrm{ISE}} \cdot \#_\mathbf{B}
\end{equation}
If $\mathrm{Score}(\mathbf{B})$ is positive, it means that matrix $\mathbf{B}$ was pruned more than the target rate, thus the fine-tuning can be done in a smaller subspace. Conversely, a negative one indicates the need for a larger parameter space. By grouping all languages for pruning in each layer, we can estimate the size of each language's intrinsic subspace in different layers respectively.

We only focus our comparison among $\mathbf{B}$ matrices because, while the $\mathbf{A}$ matrices are randomly Gaussian initialized, $\mathbf{B}$ matrices are initialized by zero in LoRA, allowing us to compare more fairly. Additionally, if we put both $\mathbf{A}$ and $\mathbf{B}$ matrices from all LoRA modules into the same pruning pool, because $\mathbf{B}$ is initialized by zero, L1 pruning will always prune all the parameters in $\mathbf{B}$ first (lead to $\mathbf{B}=\mathbf{0}$), which will cause training stability issues.

\section{Experimental Setup}
\paragraph{Dataset}
FLORES-101 \citep{goyal2021flores101} is a high-quality parallel dataset, including 3,001 sentences from English Wikipedia which are translated into 101 languages by human translators. Sentences are divided into three splits: dev (997 sentences), devtest (1,012 sentences), and test (992 sentences). Since the test set is not publicly available, we use the dev set for training and devtest set for evaluation. Languages are divided into four resource types: High (H), Medium (M), Low (L), and Very-Low (V), based on the available bitext data with English.

We first randomly selected four languages from each of the three resource types (high, medium, very-low) to form a small subset $\mathit{lang12}$ of 12 languages. We conducted comprehensive analyses and tests on $\mathit{lang12}$ to verify our proposed method. Then, we extend our method to a larger subset $\mathit{lang30}$ to measure the impact when introducing more languages. Details for $\mathit{lang12}$ and $\mathit{lang30}$ are provided in Appendix \ref{sec:ap_dataset_setting}.
\paragraph{Model Setting}
We choose M2M-124 615M \citep{goyal2021flores101} as our base model. This is a special version of M2M-100 \citep{fan2020englishcentric} extended by supplementing OPUS data to support all languages in the FLORES-101 dataset.
\paragraph{Training}
We implemented LSLo using fairseq \citep{ott-etal-2019-fairseq} based on Transformer architecture. All experiments were trained in a many-to-many setting in a single run. For full parameter fine-tuning, we trained the model for 15 epochs with a learning rate of 0.0001. For LSLo, we froze the parameters of the original model and trained for 15 epochs with a learning rate of 0.003. All models were trained on 4 RTX A6000 with automatic mixed precision. 
\paragraph{Evaluation}
We choose the results of full parameter fine-tuning as the baseline to compare with the beam size of 5. We use the dev and devtest set mentioned above as our training and test sets respectively and report $\mathit{spBLEU}$ score (SentencePiece BLEU) \citep{goyal2021flores101} with the FLORES-101 tokenizer\footnote{\url{https://github.com/facebookresearch/flores/tree/main/previous_releases/flores101}}.
\section{Results}
\subsection{Weight Learning}
\label{sec:Results_WL}
\begin{figure}[t]
  \includegraphics[width=\columnwidth]{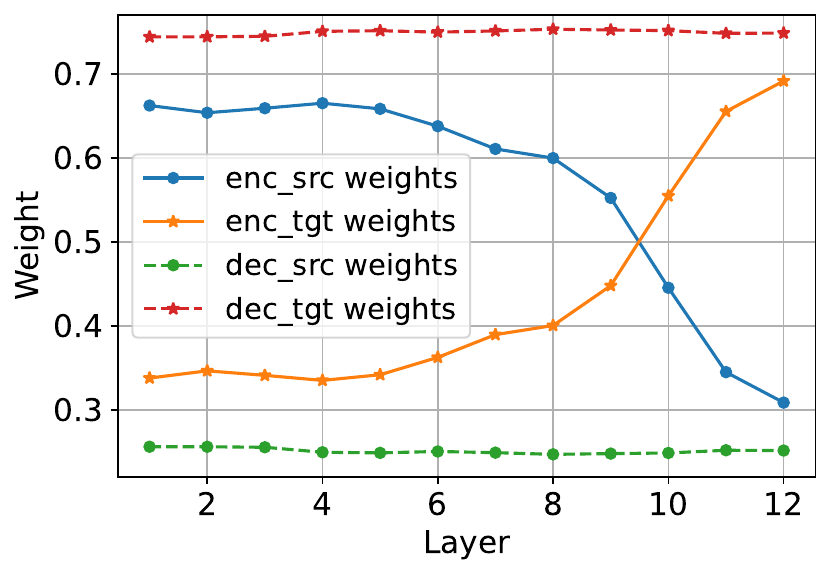}
  \caption{Source (src) and target (tgt) weights learned across layers in encoder (enc) and decoder (dec). The model's focus shifted from the source side to the target side near the top of the encoder.}
  \label{fig:weight_learning}
\end{figure}
As described in Section \ref{sec:WL}, we apply weight learning to the training data of $\mathit{lang12}$ before conducting subsequent experiments to determine whether each LSLo module should be source-indexed or target-indexed. We build both source-indexed and target-indexed LSLo modules with the same rank $r_{l_i}=8$ for all languages to all weight matrices in both encoder and decoder, including q, k, v, c-q, c-k, c-v, i.e., query, key and value matrices of attention and cross-attention respectively, and fc1, fc2, i.e., down and up matrices of MLP, respectively. In forward pass, we calculated the weighted sum of these two different indexed modules.

Figure \ref{fig:weight_learning} shows a clear tendency that the model's focus moves from the source side to the target side near the top of encoder, and in decoder, the model only focuses on target information. This is also mentioned by \citet{tenney-etal-2019-bert,pires-etal-2023-learning}, where the bottom layers of encoder only learn some lower-level representation, and the top layers capture more output-related higher-level representation. 

For the following experiments, we choose the indexed modules with the larger weights, which means the LSLo modules in the first 9 layers of encoder will be source-indexed and in other layers of encoder and decoder will be target-indexed.
\subsection{Intrinsic Subspace Estimation}
\label{sec:res_int_est}
\begin{figure}[t]
  \includegraphics[width=\columnwidth]{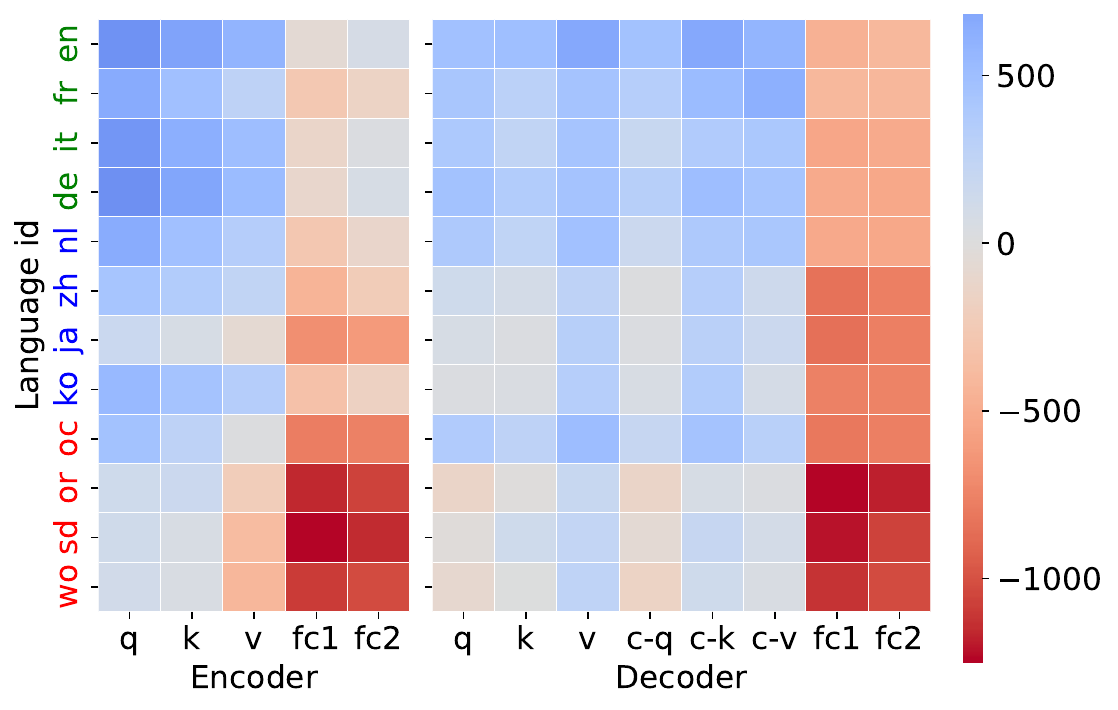}
    \caption{Illustration of the parameter space demands for each language, averaged across all layers. Color indicates the demands from low (blue) to high (red). Rows are organized by language resource type: high-resource (green), medium-resource (blue), and very-low-resource (red). Columns are organized by weight matrices in the encoder and decoder: query, key, and value matrices of attention (q, k, v) and cross-attention (c-q, c-k, c-v); down and up matrices of MLP (fc1, fc2).}
  \label{fig:intrinsic-est-avg}
\end{figure}
We performed layer-wise cross-language pruning as described in Section \ref{sec:intrinsic_est} on the training data of $\mathit{lang12}$ to estimate the space demand for each language. We added LSLo with $r_{l_i}$ for all languages to all weight matrices, allowing us to assess the parameter demands of different languages in each layer of encoder and decoder. See Appendix \ref{app:GPS} for more details of layer-wise cross-language pruning. Figure \ref{fig:intrinsic-est-avg} shows the demands calculated by Equation \ref{eq:IntrinsicEstimation} of each language, averaged across all layers. 12 languages are organized by three resource types: high-resource (green), medium-resource (blue) and very-low-resource (red).

The results indicate that the intrinsic subspace for each language is highly correlated with the resource type. Very-low-resource languages need more parameters to learn the language-specific knowledge compared to high and medium-resource ones. This suggests there is no need to use the same architecture for all languages during fine-tuning. 
We observed similar tendencies in all layers, and the details are provided in Appendix \ref{app:intrinsic_est}. 
Additionally, we also notice that compared to other languages in the same group, Dutch (nl) and Occitan (oc) require smaller parameter spaces. For Dutch (nl), it has much more bitext resources (82.4M) compared with the other three languages in the same group: Chinese (zh) (37.9M), Japanese (ja) (23.2M), Korean (ko) (7.46M). We think the resource type, which is close to high-resource languages, allows Dutch (nl) to have a smaller intrinsic subspace. For Occitan (oc), although it has only 5.11K bitext resources, it is the only language in the group that belongs to the same Language Family (Romance) as two high-resource languages, French (fr) and Italian (it). This suggests that similar languages can benefit more from cross-language learning, in line with \citet{ma-etal-2023-clustering}'s approach of integrating similar languages into a single module.

For the following experiments, we further reduce the subspace size for high and medium-resource languages by lowering their rank and applying unstructured pruning with Gradual Pruning Schedule to further explore the minimal possible intrinsic subspace.
\subsection{Main Results}
\label{subsec:MainResults}

\begin{table*}[h]
\resizebox{\textwidth}{!}{
\centering
    \begin{tabular}{llc|ccccccccc|c}
        \toprule
        & &   &   \multicolumn{9}{c|}{Language Direction} \\
        & Methods & \#Params & H2H & H2M & H2V & M2H & M2M & M2V & V2H & V2M & V2V & AVG \\ \midrule
        \multirow{2}{*}{Baselines}& Pretrain  & -     & 31.76 & 20.06 & 5.56 & 20.71 & 17.12 & 3.47 & 9.24 & 5.03 & 0.52 & 12.26 \\ 
        & Ft-all  &  615M & 29.29 & 20.46 & 12.53 & 19.28 & 17.14 & 8.95 & 15.23 & 11.02 & 6.66 & 15.43 \\ \hline
        \multirow{4}{*}{\begin{minipage}[t]{0.1\textwidth}LSLo\\+WL\end{minipage}}& 4;4;4+WL           & 15.35M & 30.15 & 20.35 & 11.76 & 19.49 & 17.04 & 8.13 & 14.58 & 10.02 & 5.66 & 15.03 \\
        & 8;8;8+WL           & 30.7M & 28.49 & 19.26 & 13.01 & 18.39 & 16.05 & 9.21 & 14.28 & 9.86 & 6.94 & 14.86 \\
        & 16;16;16+WL        & 61.4M & 25.90 & 17.82 & 14.32 & 16.86 & 14.93 & 10.57 & 13.80 & 9.71 & 8.46 & 14.55 \\
        & 64;64;64+WL        & 245.6M & 21.91 & 14.94 & \textbf{14.91} & 14.44 & 12.43 & \textbf{11.47} & 12.55 & 8.98 & 
        \textbf{9.96} & 13.40 \\ \hline
        \multirow{8}{*}{\begin{minipage}[t]{0.1\textwidth}LSLo\\+WL\\+GPS\end{minipage}}& 2;2;8+WL           & 15.3M & 31.33 & 21.07 & 13.07 & 20.16 & 17.58 & 9.3 & 15.95 & 10.89 & 7.01 & 16.05 \\
        & 2;2;8+WL+GPS(0.1)  & 15.3M & 31.37 & 21.21 & 12.90 & 20.22 & 17.63 & 9.18 & 15.93 & 10.90 & 6.96 & 16.04 \\
        & 2;2;8+WL+GPS(0.3)  & 15.3M & 31.53 & 21.33 & 12.88 & 20.32 & 17.67 & 9.18 & 15.93 & 10.84 & 6.99 & 16.08 \\
        & 2;2;8+WL+GPS(0.5)  & 15.3M & 31.76 & 21.5 & 12.96 & 20.49 & 17.94 & 9.25 & 16.08 & 10.98 & 7.1 & 16.23 \\
        & 2;2;8+WL+GPS(0.7)  & 15.3M & 32.22 & 21.81 & 12.86 & 20.92 & 18.10 & 9.22 & 16.28 & 11.12 & 6.94 & 16.38 \\
        & 2;2;8+WL+GPS(0.9)  & 15.3M & \textbf{33.13} & \textbf{22.33} & 12.93 & \textbf{21.49} & \textbf{18.58} & 9.23 & 16.59 & 11.38 & 7.04 & 16.73 \\ 
        
        & 2;2;16+WL+GPS(0.9) & 25.6M & 33.06 & 22.27 & 14.24 & 21.44 & 18.58 & 10.49 & 17.44 & 12.02 & 8.42 & 17.33 \\ 
        & 2;2;64+WL+GPS(0.9) & 86.9M & 33.02 & 22.27 & 13.96 & 21.47 & 18.56 & 10.92 & \textbf{18.67} & \textbf{12.98} & 9.48 & \textbf{17.70} \\
        \bottomrule
    \end{tabular}
}
    \caption{The spBLEU scores on $\mathit{lang12}$ organized by language resource type: High-resource (H), Medium-resource (M) and Very-low-resource (V), with the format \{H;M;V\} to show the rank we use for different languages in LSLo. WL means we follow the learned architecture of Weight Learning mentioned in Section \ref{sec:WL}. GPS($P_r$) means we use the Gradual Pruning Schedule mentioned in Section \ref{sec:GPS} for High and Medium languages with the Pruning Ratio $P_r$. Our most efficient structure (2;2;8+WL+GPS(0.9)) outperforms full parameter fine-tuning across all language directions with a much smaller number of trainable parameters \#Params.} 
    \label{tab:results_lang12} 
\end{table*}
In Table \ref{tab:results_lang12}, we report the spBLEU scores of $\mathit{lang12}$, organized by languages' resource types: High (H), Medium (M), and Very-low (V). The first column shows the experimental settings. We use the format \{H;M;V\} to show the rank in LSLo for languages with different resource types. The notation WL means we use the architecture learned from Weight Learning in Section \ref{sec:Results_WL} and GPS($P_r$) means we use the Gradual Pruning Schedule mentioned in Section \ref{sec:GPS} for high and medium-resource languages with the Pruning Ratio $P_r$. See Appendix \ref{app:GPS} for more details of GPS. We choose the zero-shot (Pretrain) and full-parameter fine-tuning (Ft-all) results as our baselines. As shown in the first two rows, although the spBLEU of very-low-resource languages improved after full parameter fine-tuning (Ft-all), high-resource languages performed poorly due to the negative interference among languages, even worse than the zero-shot results (Pretrain).

We first experimented with the same subspace size for every language but varied ranks $r\in \{4,8,16,64\}$. Results (LSLo+WL) show a trade-off between high-resource and low-resource languages, i.e., a smaller rank can alleviate the degradation of high-resource languages, e.g., 4;4;4+WL, but limits the performance of low-resource ones compared with higher rank settings, e.g., 64;64;64+WL. This indicates that sharing the same rank among languages with different resource types is suboptimal, improving low-resource performance requires a larger rank, which leads to greater degradation of high-resource performance. Although LSLo with $r=64$ achieves the best performance on very-low-resource directions, it incurs a large number of trainable parameters and sacrifices high-resource performance.

Based on the findings of Section \ref{sec:res_int_est} that high and medium-resource languages can be fine-tuned in smaller subspaces, we set a lower rank $r=2$ for high and medium-resource languages and $r=8$ for very-low-resource languages (2;2;8+WL). Compared with the setting of 8;8;8+WL, reducing parameter space for high and medium-resource languages can effectively alleviate the degradation without compromising the performance of very-low-resource directions. 

To further explore the minimal intrinsic subspace, we implemented the Gradual Pruning Schedule during fine-tuning mentioned in Section \ref{sec:GPS} for high and medium-resource languages. Based on the setting of 2;2;8+WL, we further reduce the parameter space for high and medium-resource languages by increasing $P_r$. We surprisingly find that, even after pruning 90\% of the LSLo parameters for high and medium-resource languages (2;2;8+WL+GPS(0.9)),  our method still achieves a 1.3 spBLEU improvement over the full parameter fine-tuning baseline, with only 2.5\% trainable parameters. Furthermore, the degradation in high-resource languages has also been solved, with H2H performance improved from a decline of -2.47 spBLEU to an increase of +1.37 spBLEU. This suggests that language-specific fine-tuning for high and medium-resource languages actually occurs within tiny subspaces. Therefore, we can save more space for low-resource language learning. Simply increasing the rank for very-low-resource languages to 64 (2;2;64+WL+GPS(0.9)) can achieve a 2.26 spBLEU improvement and is more parameter-efficient.
\begin{table*}[h]
\resizebox{\textwidth}{!}{
\centering
    \begin{tabular}{lc|cccccccccccccccc|c}
        \toprule
          &   &   \multicolumn{16}{c|}{Language Direction} \\
        Methods & \#Params & H2H & H2M & H2L & H2V & M2H & M2M & M2L & M2V & L2H & L2M & L2L & L2V & V2H & V2M & V2L & V2V & AVG \\ \midrule
        Pretrain & - & 28.93 & 20.77 & 6.29 & 3.60 & 22.94 & 17.26 & 4.82 & 2.73 & 11.28 & 8.01 & 3.03 & 1.51 & 7.04 & 4.34 & 1.68 & 0.55 & 9.53 \\
        Ft-all & 615M & 24.48 & 19.80 & 9.76 & 7.94 & 19.44 & 16.72 & 8.55 & 6.72 & 12.53 & 11.17 & 6.76 & 5.15 & 11.11 & 9.72 & 6.05 & 4.04 & 11.61 \\
        8;8;8;8+WL & 76.7M & 22.94 & 17.91 & \textbf{11.15} & \textbf{10.24} & 18.07 & 15.00 & \textbf{9.64} & 8.74 & 12.33 & 10.46 & 8.15 & 7.34 & 11.07 & 9.30 & 7.47 & 6.11 & 11.83 \\
        16;16;16;16+WL & 153.4M & 19.58 & 15.29 & 11.08 & 10.47 & 15.53 & 12.95 & \textbf{9.64} & \textbf{9.10} & 11.18 & 9.56 & \textbf{8.29} & \textbf{7.83} & 10.10 & 8.59 & \textbf{7.62} & \textbf{6.74} & 10.98 \\
        2;2;8;8+WL+GPS(0.9) & 46M & \textbf{29.92} & \textbf{22.90} & 11.11 & 10.06 & \textbf{23.60} & \textbf{19.20} & 9.53 & 8.61 & \textbf{15.34} & \textbf{12.70} & 8.05 & 7.25 & \textbf{13.75} & \textbf{11.31} & 7.37 & 6.11 & \textbf{13.86} \\
        \bottomrule
    \end{tabular}
}
    \caption{The spBLEU scores on $\mathit{lang30}$ organized by languages' resource type: High-resource (H), Medium-resource (M), Low-resource (L) and Very-low-resource (V), with the format \{H;M;L;V\} to show the rank we use for different languages in LSLo. Our most efficient structure (2;2;8;8+WL+GPS(0.9)) outperforms full parameter fine-tuning, demonstrating the effectiveness and scalability of our proposed method.}
    \label{tab:results_lang30} 
\end{table*}

We also expand our experiments to 30 languages $\mathit{lang30}$ in Table \ref{tab:results_lang30} to assess our method's scalability. Languages are divided into four resource types: High (H), Medium (M), Low (L), and Very-low (V). Similar to Table \ref{tab:lang12}, we use the format \{H;M;L;V\} to represent the rank setting in LSLo. Although the number of trainable parameters increases with the additional introduction of language-specific modules, our method (2;2;8;8+WL+GPS(0.9)) still achieved a 2.25 spBLEU improvement over full parameter fine-tuning with only 7\% trainable parameters. This demonstrates our method's potential to support hundreds of languages while still keeping the number of trainable parameters near the original model.
\section{Analysis and Discussion}
\subsection{Is Weight Learning Effective?}
\begin{table*}[t]
\resizebox{\textwidth}{!}{
\centering
    \begin{tabular}{lc|ccccccccc|c}
        \toprule
        &   &   \multicolumn{9}{c|}{Language Direction} \\
        Methods & \#Params & H2H & H2M & H2V & M2H & M2M & M2V & V2H & V2M & V2V & AVG \\ \midrule
        Pre-trained            & -     & 31.76 & 20.06 & 5.56 & 20.71 & 17.12 & 3.47 & 9.24 & 5.03 & 0.52 & 12.26 \\ 
        Ft-all              &  615M & 29.29 & 20.46 & 12.53 & 19.28 & 17.14  & 8.95 & 15.23 & 11.02 & 6.66 & 15.43 \\
        2;2;8+WL+GPS(0.9)  & 15.3M & \textbf{33.13} & 22.33 & 12.93 & \textbf{21.49} & 18.58 & \textbf{9.23} & \textbf{16.59} & \textbf{11.38} & \textbf{7.04} & \textbf{16.73} \\
        2;2;8+SRC+GPS(0.9) & 15.3M & 33.06 & \textbf{22.40} & 12.42 & 21.41 & \textbf{18.59} & 8.76 & 16.41 & 11.24 & 6.59 &  16.52 \\
        2;2;8+TGT+GPS(0.9) & 15.3M & 32.97 & 22.34 & \textbf{13.05} & 21.40 & 18.53 & \textbf{9.23} & 11.91 & 7.69 & 5.05 & 15.52 \\
        \bottomrule
    \end{tabular}
}
    \caption{The spBLEU scores of different index strategies on $\mathit{lang12}$.} 
    \label{tab:WL_Analysis} 
\end{table*}
In this section, we analyze the improvements brought by the structure learned via Weight Learning from Section \ref{sec:Results_WL}. Given the results in Figure \ref{fig:weight_learning} that the decoder always focuses on the target side information, we concentrate on comparing different encoder settings. We compared three different encoder settings in Table \ref{tab:WL_Analysis}: (1) Weight Learning (WL) as described in Section \ref{sec:Results_WL}, where LSLo modules in the first 9 layers of encoder are source-indexed and in the last 3 layers are target-indexed; (2) Source Encoder (SRC), where all LSLo modules in encoder are source-indexed; (3) Target Encoder (TGT), where all LSLo modules in encoder are target-indexed. We found that the structure selected through Weight Learning (2;2;8+WL+GPS(0.9) exhibited better overall performance, especially for very-low-resource languages.
\subsection{What Causes the Degradation of High-resource Languages?}
\begin{figure}[h]
    \begin{subfigure}{\columnwidth}
        \includegraphics[width=\columnwidth]{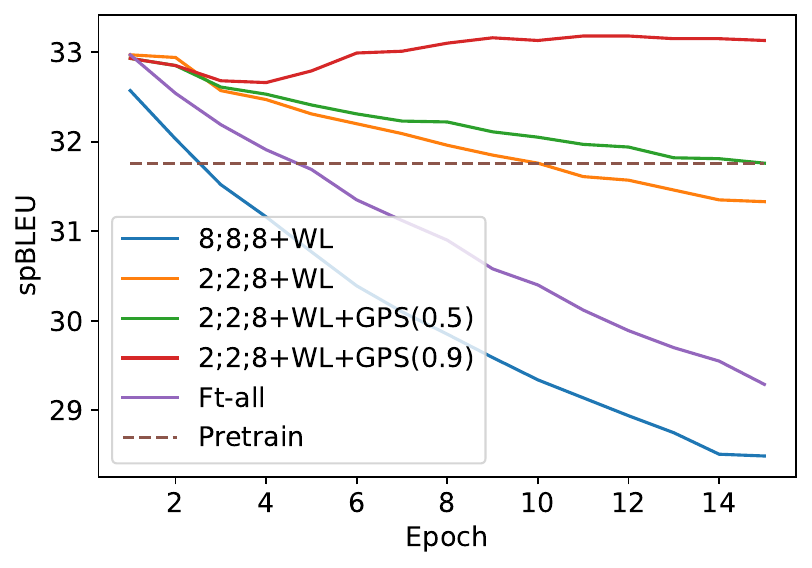}
        \caption{H2H Performance per epoch}
        \label{fig:why_degeneration_H2H}
    \end{subfigure}
    \begin{subfigure}{\columnwidth}
        \includegraphics[width=\columnwidth]{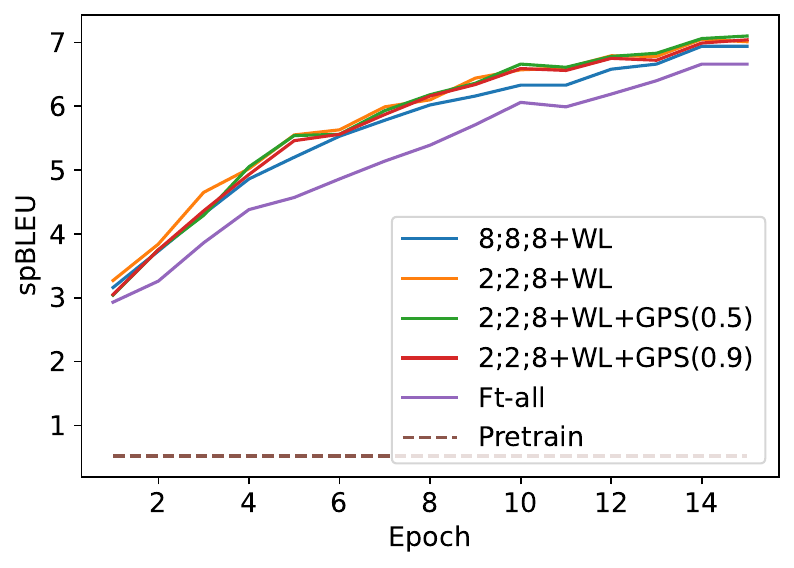}
        \caption{V2V Performance per epoch}
    \end{subfigure}
    \caption{We examined the performance of H2H and V2V directions per epoch. H2H performance declined during training.}
    \label{fig:why_degeneration}
\end{figure}
In previous experiments of Section \ref{subsec:MainResults}, we discovered that merely isolating each language representation into different parameter spaces via LSLo did not mitigate the performance degradation of high-resource languages. This indicates that the trading or competing language representation might not be the only factor causing the decline.

We examined the spBLEU of H2H and V2V directions per epoch, as shown in Figure \ref{fig:why_degeneration}. We observed that the spBLEU of low-resource languages continuously improved during training, whereas high-resource languages' performance increased in the first epoch and then gradually declined. This suggests the pre-trained model has already acquired substantial knowledge of high-resource languages, making their subspace smaller compared to low-resource ones. When allocating same-sized parameter spaces for languages of different resource types, high-resource languages are more susceptible to over-fitting, which contributes to an over-fitting phenomenon leading to degradation. 

This also explains why reducing the trainable parameter of high-resource languages can achieve better performance. As shown in Figure \ref{fig:why_degeneration_H2H}, over-fitting is mitigated by continuously reducing the subspace size (2;2;8+WL+GPS(0.9)), without compromising the performance of low-resource languages.
\subsection{Where Should We Apply LSLo to?}
\begin{table*}[t]
\resizebox{\textwidth}{!}{
\centering
    \begin{tabular}{llc|ccccccccc|c}
        \toprule
        &  &   &   \multicolumn{9}{c|}{Language Direction} \\
        & Methods & \#Params & H2H & H2M & H2V & M2H & M2M & M2V & V2H & V2M & V2V & AVG \\ \midrule
        \multirow{2}{*}{Baselines} & Pre-trained            & -     & 31.76 & 20.06 & 5.56 & 20.71 & 17.12 & 3.47 & 9.24 & 5.03 & 0.52 & 12.26 \\ 
        & Ft-all              &  615M & 29.29 & 20.46 & 12.53 & 19.28 & 17.14  & 8.95 & 15.23 & 11.02 & 6.66 & 15.43 \\ \hline
        \multirow{2}{*}{All} & 2;2;8+WL+GPS(0.9)  & 15.3M & 33.13 & \textbf{22.33} & 12.93 & 21.49 & 18.58 & 9.23 & 16.59 & 11.38 & 7.04 & 16.73 \\
        & 2;2;16+WL+GPS(0.9) & 25.6M & 33.06 & 22.27 & 14.24 & 21.44 & 18.58 & 10.49 & 17.44 & 12.02 & 8.42 & 17.33 \\ \hline
        \multirow{2}{*}{Only attn} & 2;2;8+WL+GPS(0.9) & 10.6M & 33.16 & 22.32 & 10.83 & 21.47 & 18.59 & 7.37 & 15.12 & 10.11 & 4.65 & 15.70 \\
        & 2;2;16+WL+GPS(0.9) & 17.7M & 33.16 & 22.25 & 12.00 & 21.45 & \textbf{18.61} & 8.34 & 15.91 & 10.76 & 5.88 & 16.24 \\ \hline        
        \multirow{3}{*}{Only fc} & 2;2;16+WL+GPS(0.9) & 7.7M & 33.29 & 22.19 & 12.64 & \textbf{21.60} & 18.56 & 8.83 & 17.33 & 11.65 & 6.90 & 16.76 \\
        & 2;2;32+WL+GPS(0.9) & 14.1M & 33.24 & 22.31 & 14.11 & 21.50 & 18.46 & 10.12 & 18.19 & 12.47 & 8.19 & 17.44 \\
        & 2;2;64+WL+GPS(0.9) & 26.7M & \textbf{33.27} & 22.26 & \textbf{14.86} & 21.59 & 18.48 & \textbf{10.95} & \textbf{18.97} & \textbf{12.97} & \textbf{9.79} & \textbf{17.91} \\
        \bottomrule
    \end{tabular}
}
    \caption{We compare the performance on $\mathit{lang12}$ of adding LSLo to all modules (with *) versus only adding it to fully connected layers (Only fc) and only adding it to attention modules (Only attn). We found that, given a similar parameter budget, adding LSLo to fc1 and fc2 results in better performance.} 
    \label{tab:fc} 
\end{table*}
\label{sec:res_languagespecpruning}
In this section, we want to discuss which weight matrices in the Transformer architecture are more crucial for LSLo. Similar to Section \ref{sec:intrinsic_est}, we employ language-specific pruning on the training data of $\mathit{lang12}$ to measure the demands for different weight matrices using Equation \ref{eq:IntrinsicEstimation}. Specifically, we add LSLos with a rank of 8 to all possible weight matrices and group the $B$ matrix from all LoRA modules for each language into respective pruning groups. See Appendix \ref{app:GPS} for more details of language-specific pruning. In this setting, we aim to examine which weight matrices are more important for different languages. The results averaged across all 12 languages are shown in Figure \ref{fig:lang-pruning-avg}. Further details for each language respectively are shown in Appendix \ref{app:lspruning}. We observed a clear trend across all 12 languages: fc1 and fc2 play a more important role in both encoder and decoder compared to other weight matrices. This is in line with the observation by \citet{geva-etal-2021-transformer} that feed-forward layers in Transformer architecture function as key-value memories for refining the final output, thus more crucial than other weight matrices. 

In Table \ref{tab:fc}, we compared the results on $\mathit{lang12}$ of applying LSLo to all weight matrices (All) versus only applying it to fc1 and fc2 (Only fc) and only applying it to attention modules including cross-attention (Only attn), given a similar parameter budget. We found that applying LSLo only to fc1 and fc2 consistently yields better results. This suggests that, under a limited parameter budget, concentrating parameters in the feed-forward layers are more effective than distributing them across all possible weight matrices.

\begin{figure}[t]
  \includegraphics[width=\columnwidth]{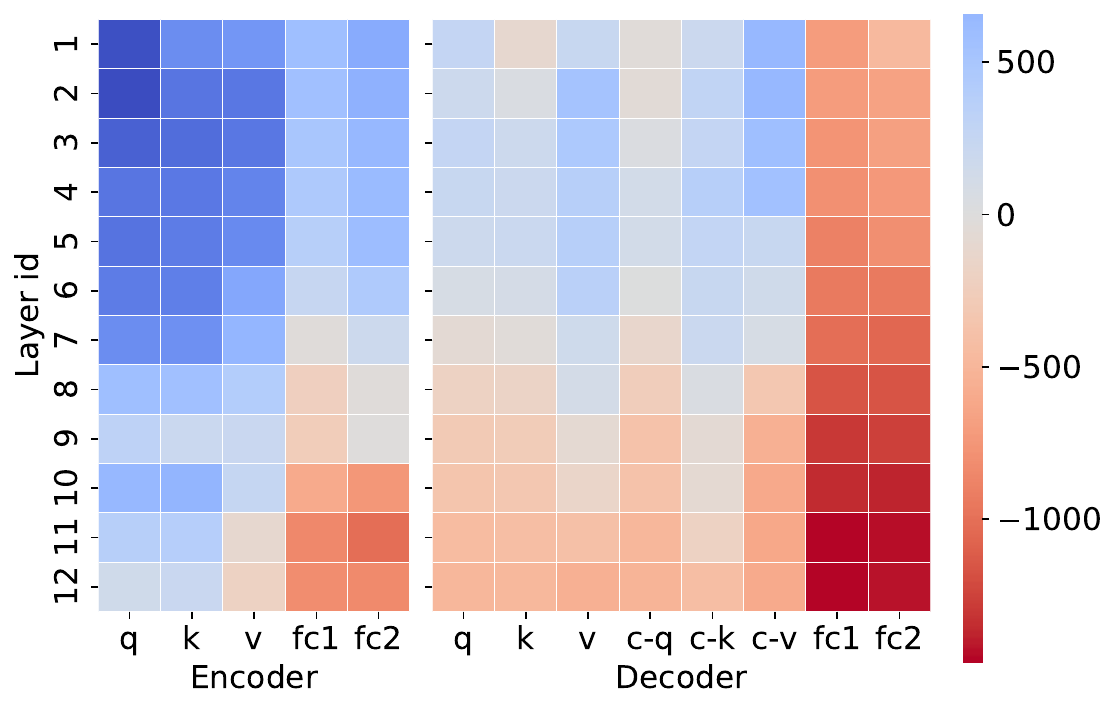}
  \caption{Illustration of the parameter space demands for each weight matrix, averaged across all languages. Color indicates the demands from low (blue) to high (red). Columns are organized by weight matrices in the encoder and decoder: query, key, and value matrices of attention (q, k, v) and cross-attention (c-q, c-k, c-v); down and up matrices of MLP (fc1, fc2).}
  \label{fig:lang-pruning-avg}
\end{figure}
\section{Related work}
\paragraph{Intrinsic Subspace}
Intrinsic Subspace is the minimal parameter subspace required for models to learn new tasks. \citet{li2018measuring} first showed that intrinsic structures exist in deep neural networks through random subspace projection. \citet{aghajanyan-etal-2021-intrinsic} further used this concept to explain the fine-tuning of Pre-trained Models. Following their works, \citet{qin2022exploring} found a universal task subspace that only includes hundreds of parameters through prompt-tuning \citep{NEURIPS2020_1457c0d6,li-liang-2021-prefix,liu-etal-2022-p}, and \citet{zhang-etal-2023-fine} observe outlier dimensions during fine-tuning. However, their experiments do not include natural language generation (NLG) tasks. To bridge this gap, our work focuses on Multilingual Neural Machine Translation, a particularly challenging NLG task.
\paragraph{Low-Rank Adaptation (LoRA)} 
LoRA \citep{hu2021lora} employs the product of two low-rank matrices to replace the original parameter matrix for fine-tuning. This method is parameter-efficient and widely used in Large Language Models. Recent works \citep{zhang2023adalora,kopiczko2024vera} have focused on how to further enhance the efficiency of LoRA. \citet{zhang2023adalora} modeled LoRA in the form of singular value decomposition and improved efficiency by pruning less important singular values. \citet{kopiczko2024vera} reduced trainable parameters of LoRA by only leaning scaling vectors during training, fixed low-rank matrices are randomly initialized and shared for each layer. Inspired by these works, we propose LSLo, a LoRA based method, to model the intrinsic subspace of language-specific learning.
\paragraph{Language-specific Learning}
Multilingual models suffer from the negative interaction among languages \citep{duh-etal-2012-learning,chen2023pareto,huang-etal-2023-towards}. Introducing language-specific structures is a common strategy to address this issue. \citet{sachan-neubig-2018-parameter, escolano-etal-2021-multilingual, pires-etal-2023-learning} built language-specific encoder or decoder layers. Despite its effectiveness, a large number of trainable parameters are required for such architecture. Another line of work \citep{lin-etal-2021-learning,Wang_Zhang_2022,he-etal-2023-gradient} tried to extract sub-networks by first fine-tuning on all language pairs separately and then jointly training these sub-networks. However, the number for fine-tuning will increase quadratically with the number of languages, consuming significant computational resources. In this work, we propose a parameter-efficient method that maximizes the utilization of the substantial knowledge learned by Pre-trained Multilingual Models to improve the performance of all language pairs.
\section{Conclusion}
In this work, we studied the imbalance size distribution of intrinsic language-specific subspaces in a Pre-trained Multilingual Model. We modeled the intrinsic language-specific subspaces using LSLo. We further proposed an intrinsic subspace estimation method and found that the size of the intrinsic subspace for each language is highly correlated with its resource type. The required subspace size for higher-resource languages is much smaller than for lower-resource ones. Therefore, there is no need to set the same parameter budget for all languages when fine-tuning multilingual models. By fine-tuning languages in their respective intrinsic subspaces with different sizes using LSLo, we achieved significant improvements compared to full parameter fine-tuning while greatly reducing the number of trainable parameters. We also proposed methods to search for the optimal placement of LSLo. We showed that the model completes the transformation from the source side to the target side in the top layers of the encoder and that placing the LSLo module in the fully connected layers is most effective in the Transformer architecture.
\section*{Limitations}
Despite the insights gained from our work, our research still has some limitations.

During the experiments, we categorized languages based on resource types, which is still a relatively coarse classification. We believe that setting individual ranks and pruning ratios for each language could further improve performance and efficiency. Although we did not conduct experiments for all the languages due to time constraints, our proposed optimal architecture search methods can support analysis for each language respectively.

Our experiments only used M2M124-615M Model. We believe that introducing more languages and larger-scale models would yield more interesting findings. However, due to resource and time constraints, it is challenging to use large language models for many-to-many training and conduct comprehensive analysis.

\bibliography{custom}
\appendix

\clearpage
\section{Dataset Setting}
The details of $\mathit{lang12}$ and $\mathit{lang30}$ are reported in Table \ref{tab:lang12} and Table \ref{tab:lang30}. We follow the resource type classification from FLORES-101 \citep{goyal2021flores101} based on available Bitext data through English (Bitext w/En). We use the language code of M2M-124 model. The Language Family information and available Bitext data through English are all from FLORES-101.
\label{sec:ap_dataset_setting}
\begin{table}[t]
\resizebox{\columnwidth}{!}{
\centering
\begin{tabular}{lllll}
    \toprule
    Resource Type & Language & Code & Family & Bitext w/ En  \\
    \midrule
    \multirow{4}{*}{High} & English & en & Germanic & - \\
     & French & fr & Romance & 289M \\
     & German & de & Germanic & 216M \\
     & Italian & it & Romance & 116M \\
    \midrule
    \multirow{4}{*}{Medium} & Chinese & zh & Sino-Tibetan & 37.9M \\
     & Dutch & nl & Germanic & 82.4M \\
     & Japanese & ja & Japonic & 23.2M \\
     & Korean & ko & Koreanic & 7.46M \\
    \midrule
    \multirow{4}{*}{Very Low} & Occitan & oc & Romance & 5.11K \\
    & Oriya & or & Indo-Aryan & 5K \\
    & Sindhi & sd & Indo-Aryan & 21.8K \\
    & Wolof & wo & Nilotic+Other AC & 86.9K \\
    \bottomrule
\end{tabular}
}
\caption{Details for each language in $\mathit{lang12}$.}
\label{tab:lang12}
\end{table}

\begin{table}[t]
\resizebox{0.95\columnwidth}{!}{
\centering
\begin{tabular}{lllll}
    \toprule
    Resource Type & Language & Code & Family & Bitext w/ En  \\
    \midrule
    \multirow{7}{*}{High} & English & en & Germanic & - \\
     & French & fr & Romance & 289M \\
     & German & de & Germanic & 216M \\
     & Italian & it & Romance & 116M \\
     & Portuguese & pt & Romance & 137M \\
     & Russian & ru & Balto-Slavic & 127M \\
     & Spanish & es & Romance & 315M \\ 
    \midrule
    \multirow{9}{*}{Medium} 
     & Arabic & ar & Afro-Asiatic & 25.2M \\
     & Chinese & zh & Sino-Tibetan & 37.9M \\
     & Dutch & nl & Germanic & 82.4M \\
     & Hebrew & he & Afro-Asiatic & 6.64M \\
     & Hindi & hi & Indo-Aryan & 3.3M \\
     & Japanese & ja & Japonic & 23.2M \\
     & Korean & ko & Koreanic & 7.46M \\
     & Maltese & mt & Afro-Asiatic & 5.82M \\
     & Norwegian & no & Germanic & 10.9M \\
    \midrule
    \multirow{8}{*}{Low} & Afrikaans & af & Germanic & 570K \\
    & Amharic & am & Afro-Asiatic & 339K \\
    & Armenian & hy & Other IE & 977K \\
    & Hausa & ha & Afro-Asiatic & 335K \\
    & Nyanja & ny & Bantu & 932K \\
    & Shona & sn & Bantu & 877K \\
    & Yoruba & yo & Nilotic+Other AC & 171K \\
    & Zulu & zu & Bantu & 123K \\
    \midrule
    \multirow{6}{*}{Very Low} 
    & Fula & ff & Nilotic+Other AC & 71K \\
    & Kamba & kam & Bantu & 50K \\
    & Occitan & oc & Romance & 5.11K \\
    & Oriya & or & Indo-Aryan & 5K \\
    & Sindhi & sd & Indo-Aryan & 21.8K \\
    & Wolof & wo & Nilotic+Other AC & 86.9K \\
    \bottomrule
\end{tabular}
}
\caption{Details for each language in $\mathit{lang30}$.}
\label{tab:lang30}
\end{table}
\section{Gradual Pruning Schedule}
\label{app:GPS}
In Table \ref{tab:GPS-Settings}, we show the settings of Gradual Pruning Schedule in different experiments, where ISE denotes Intrinsic Subspace Estimation mentioned in Section \ref{sec:res_int_est}, LSP denotes Language-specific Pruning mentioned in Section \ref{sec:res_languagespecpruning} and LSLo denotes the Language-specific LoRA in Section \ref{subsec:MainResults}. We empirically set $P_\mathrm{ISE}=0.7$ and $P_\mathrm{LSP}=0.7$. If $P$ is too small, it may not effectively demonstrate the differences between different languages. Conversely, if $P$ is too large, it can lead to overly aggressive pruning, making some LoRA modules entirely pruned and causing training issues. Therefore, based on the model size, we think $0.7$ is a reasonable number for ISE and LSP. For $P_\mathrm{LSLo}$, as shown in Table \ref{tab:results_lang12}, we tried different values to exhaustively explore the possible minimal intrinsic subspace.
\begin{table}[h]
\resizebox{\columnwidth}{!}{
    \centering
    \begin{tabular}{lcccc}
        \toprule
        Experiments & $P$ & E & k & T \\
        \midrule
        ISE & 0.7 & 2 & 8 & 15 \\
        LSP & 0.7 & 2 & 8 & 15 \\
        LSLo & \{0.1,0.3,0.5,0.7,0.9\} & 2 & 8 & 15 \\
        \bottomrule
    \end{tabular}
}
\caption{Settings of Gradual Pruning Schedule in different experiments.}
\label{tab:GPS-Settings}
\end{table}

\begin{table*}[t]
\resizebox{\textwidth}{!}{
\centering
    \begin{tabular}{lc|ccccccccc|c}
        \toprule
          &   &   \multicolumn{9}{c|}{Language Direction} \\
        Pruning Strategy & \#Params & H2H & H2M & H2V & M2H & M2M & M2V & V2H & V2M & V2V & AVG \\ \midrule
        Pre-trained            & -     & 31.76 & 20.06 & 5.56 & 20.71 & 17.12 & 3.47 & 9.24 & 5.03 & 0.52 & 12.26 \\ 
        Ft-all              &  615M & 29.29 & 20.46 & 12.53 & 19.28 & 17.14  & 8.95 & 15.23 & 11.02 & 6.66 & 15.43 \\
        E=2,K=8  & 15.3M & 33.13 & 22.33 & 12.93 & 21.49 & 18.58 & 9.23 & 16.59 & 11.38 & 7.04 & 16.73 \\
        E=5,K=8  & 15.3M & 33.05 & 22.23 & \textbf{13.10} & 21.51 & 18.50 & \textbf{9.36} & 16.79 & 11.46 & \textbf{7.16} & 16.79 \\
        E=7,K=8  & 15.3M & 32.96 & 22.14 & 12.90 & 21.38 & 18.43 & 9.24 & 16.50 & 11.28 & 7.02 & 16.64 \\
        E=2,K=2  & 15.3M & \textbf{33.16} & \textbf{22.46} & 13.04 & 21.49 & \textbf{18.61} & 9.33 & \textbf{16.84} & \textbf{11.52} & 7.13 & \textbf{16.83} \\
        E=7,K=2  & 15.3M & 33.13 & 22.31 & 12.88 & \textbf{21.58} & 18.56 & 9.22 & 16.71 & 11.38 & 6.96 & 16.74 \\
        \bottomrule
    \end{tabular}
}
    \caption{2;2;8+WL+GPS(0.9) with different pruning strategies on $\mathit{lang12}$ (start from $E$ epoch, end at $E+k$ epoch, $T=15$).}
    \label{tab:EkT} 
\end{table*}
We set the same $E=2$, $k=8$, $T=15$ based on the following two intuitions: (1) Given the high pruning ratio for high-resource languages, more training epochs should follow after pruning; (2) Pruning should last for more epochs to ensure the stability of the training process. We want to emphasize that the purpose of using Gradual Pruning Schedule is stability in training instead of searching for good parameters under a specific hyperparameter setting. As shown in the following Table \ref{tab:EkT}, we also checked different start epoch $E$ and duration $k$. Except for the unreasonable setting of $E=7$,$k=8$ where there is no enough training after pruning, other settings achieve better performance than the score we reported but no significant differences (max to $0.1$). Considering the space limitation and the limited improvement ($+0.01\sim+0.1$) compared with our proposed method (more than 1), we simply choose the empirical setting in our paper for brevity. 

\begin{table}[t]
\resizebox{\columnwidth}{!}{
    \centering
    \begin{tabular}{lcc}
        \toprule
        Module & Correlation Coefficients & P-value \\
        \midrule
        q & 0.67 & 0.023 \\
        k & 0.64 & 0.032 \\
        v & 0.60 & 0.050 \\
        fc1 & 0.65 & 0.029 \\
        fc2 & 0.64 & 0.033 \\
        AVG & 0.64 & 0.033 \\
        \bottomrule
    \end{tabular}
}
\caption{The correlation coefficients in Encoder.}
\label{tab:ccEncoder}
\end{table}
\begin{table}[t]
\resizebox{\columnwidth}{!}{
    \centering
    \begin{tabular}{lcc}
        \toprule
        Module & Correlation Coefficients & P-value \\
        \midrule
        q & 0.65 & 0.029 \\
        k & 0.58 & 0.059 \\
        v & 0.47 & 0.143 \\
        c-q & 0.70 & 0.017 \\
        c-k & 0.59 & 0.052 \\
        c-v & 0.80 & 0.003 \\
        fc1 & 0.72 & 0.012 \\
        fc2 & 0.74 & 0.008 \\
        AVG & 0.66 & 0.041 \\
        \bottomrule
    \end{tabular}
}
\caption{The correlation coefficients in Encoder.}
\label{tab:ccDecoder}
\end{table}
\section{Intrinsic Subspace Estimation}
\label{app:intrinsic_est}
\begin{figure*}[t]
    \begin{subfigure}{0.66\columnwidth}
        \includegraphics[width=\columnwidth]{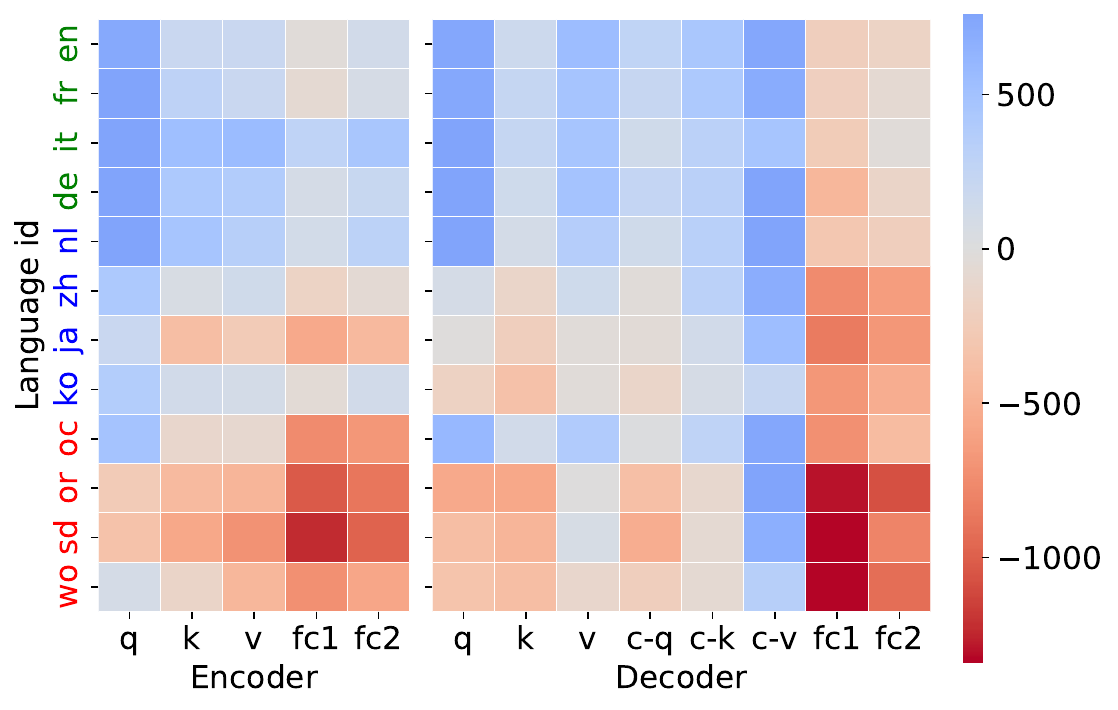}
        \caption{Layer 1}
    \end{subfigure}
    \begin{subfigure}{0.66\columnwidth}
        \includegraphics[width=\columnwidth]{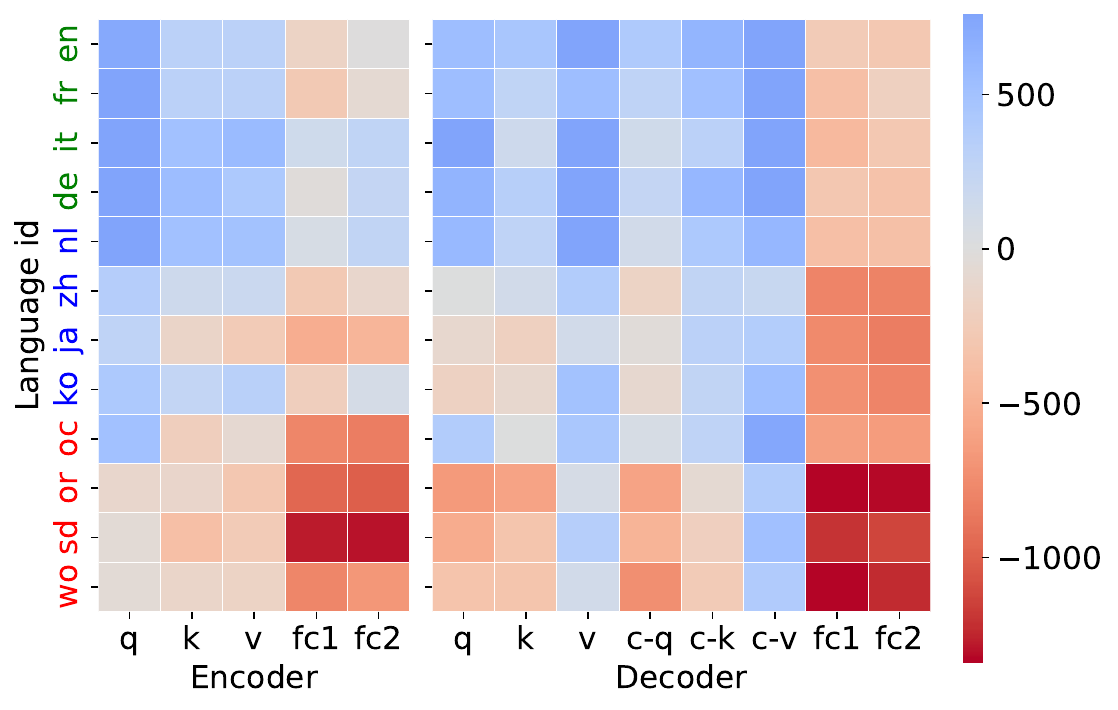}
        \caption{Layer 2}
    \end{subfigure}
    \begin{subfigure}{0.66\columnwidth}
        \includegraphics[width=\columnwidth]{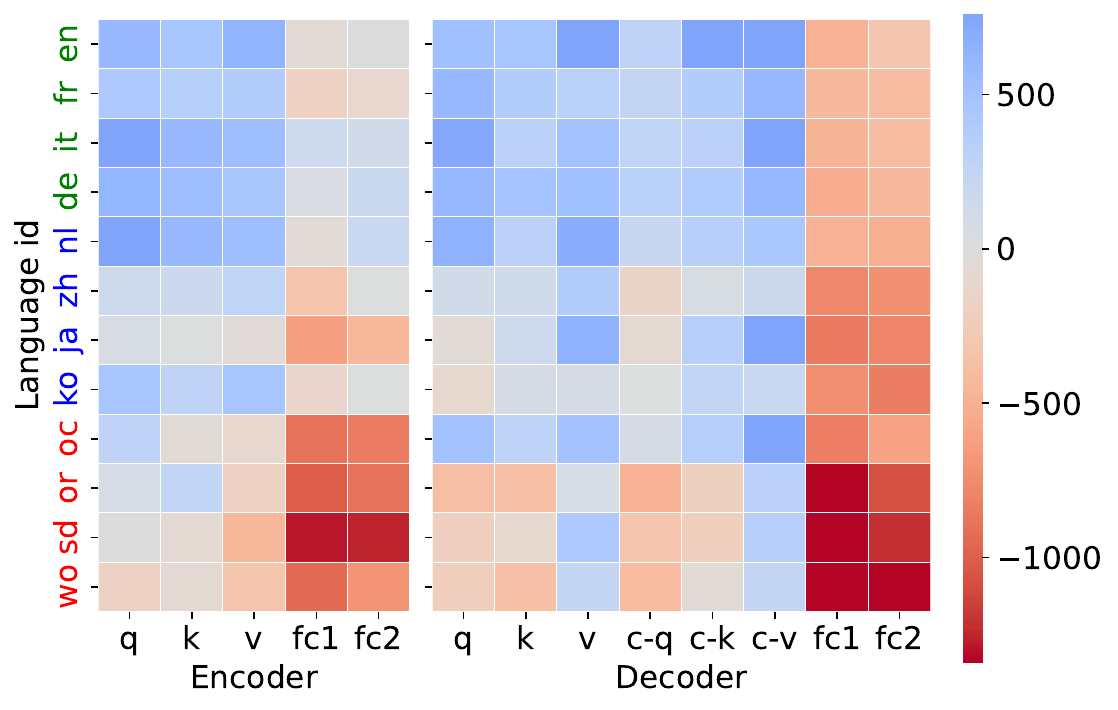}
        \caption{Layer 3}
    \end{subfigure}
    \begin{subfigure}{0.66\columnwidth}
        \includegraphics[width=\columnwidth]{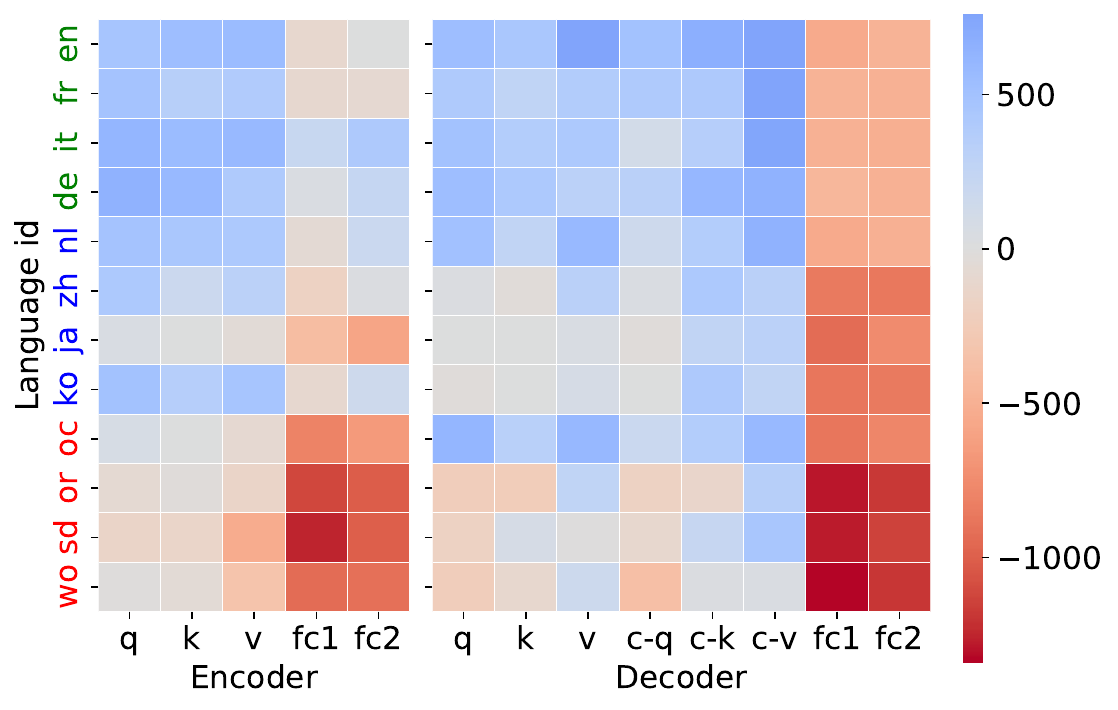}
        \caption{Layer 4}
    \end{subfigure}
    \begin{subfigure}{0.66\columnwidth}
        \includegraphics[width=\columnwidth]{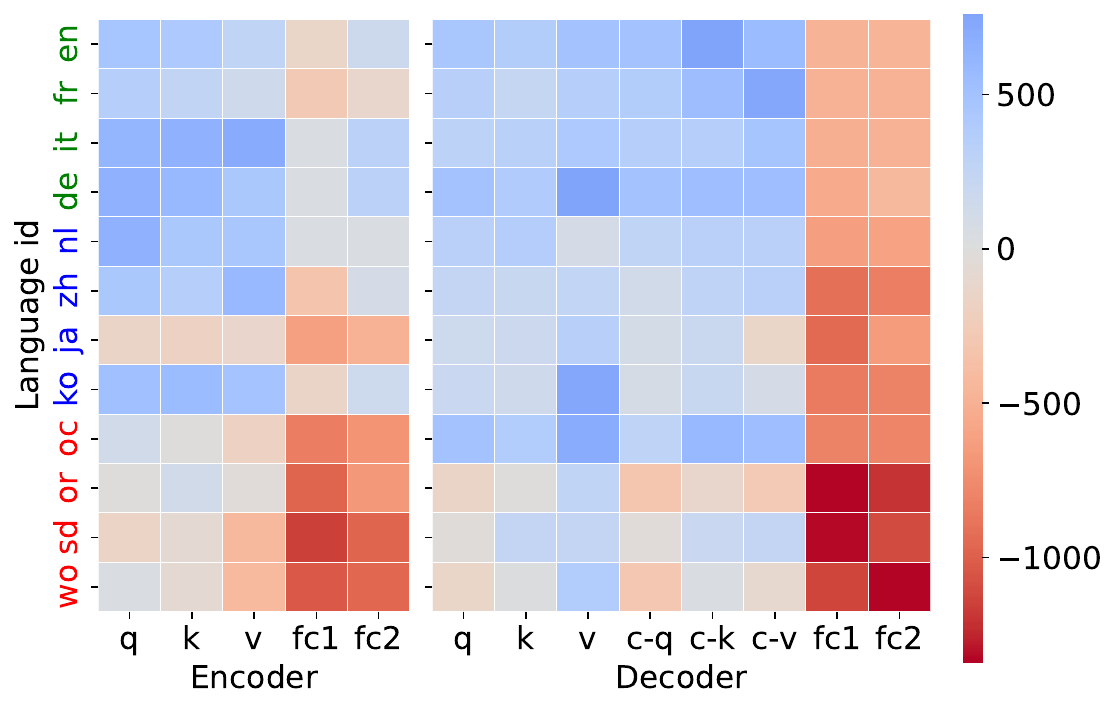}
        \caption{Layer 5}
    \end{subfigure}
    \begin{subfigure}{0.66\columnwidth}
        \includegraphics[width=\columnwidth]{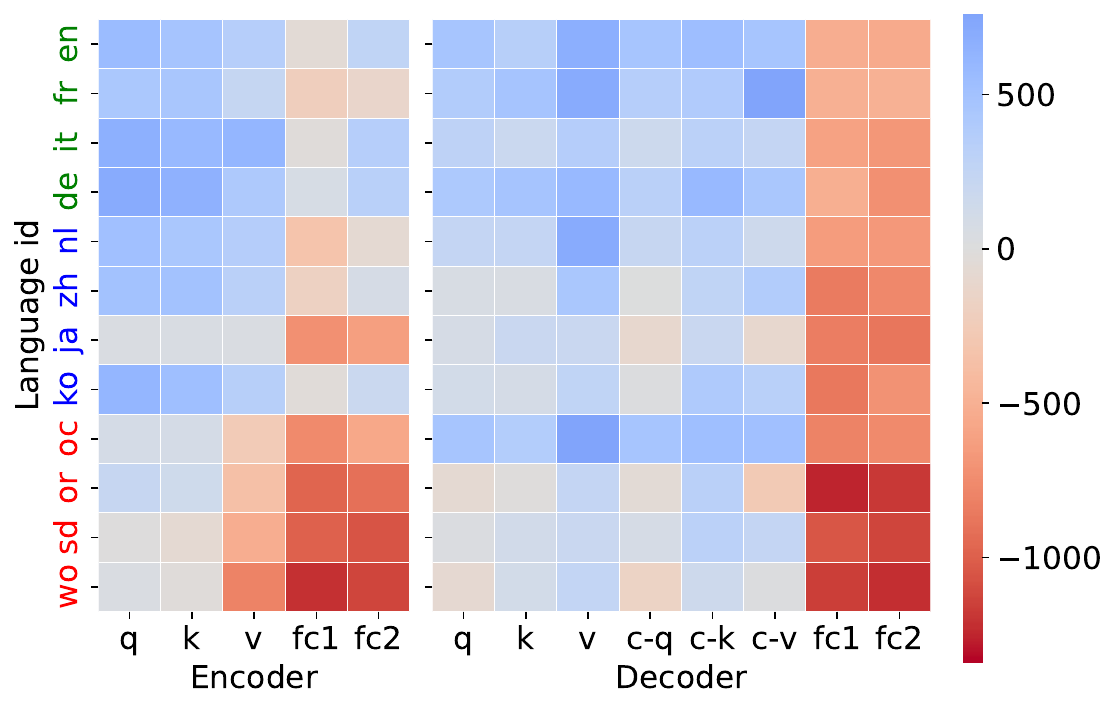}
        \caption{Layer 6}
    \end{subfigure}
    \hfill
    \begin{subfigure}{0.66\columnwidth}
        \includegraphics[width=\columnwidth]{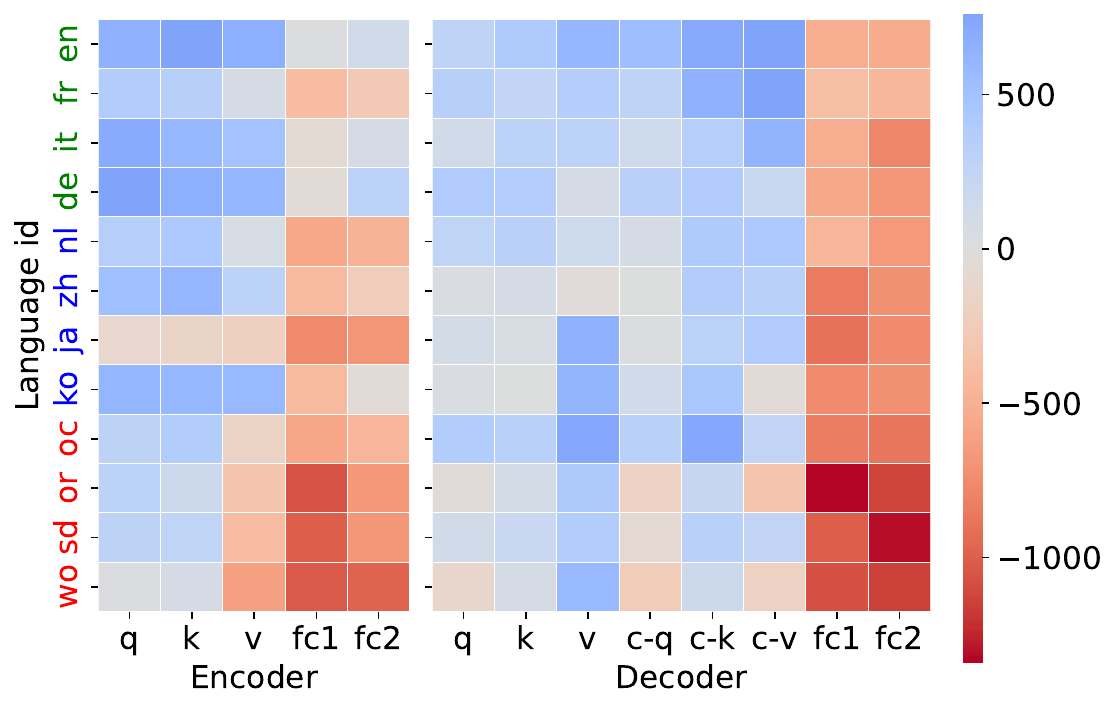}
        \caption{Layer 7}
    \end{subfigure}
    \begin{subfigure}{0.66\columnwidth}
        \includegraphics[width=\columnwidth]{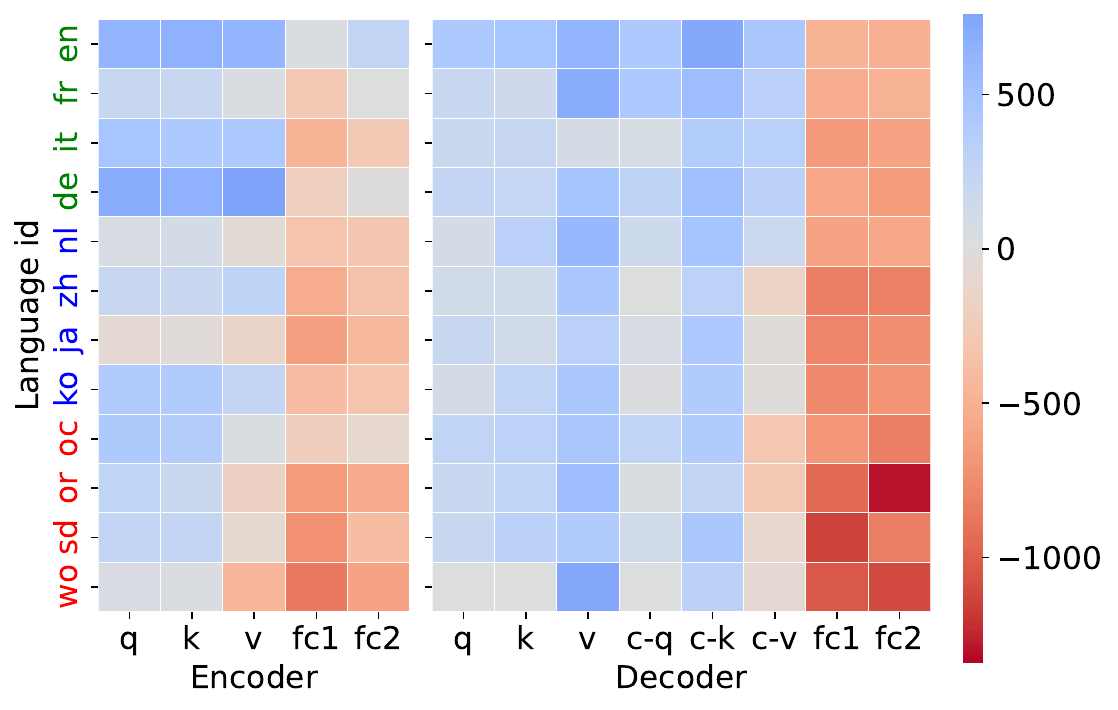}
        \caption{Layer 8}
    \end{subfigure}
    \begin{subfigure}{0.66\columnwidth}
        \includegraphics[width=\columnwidth]{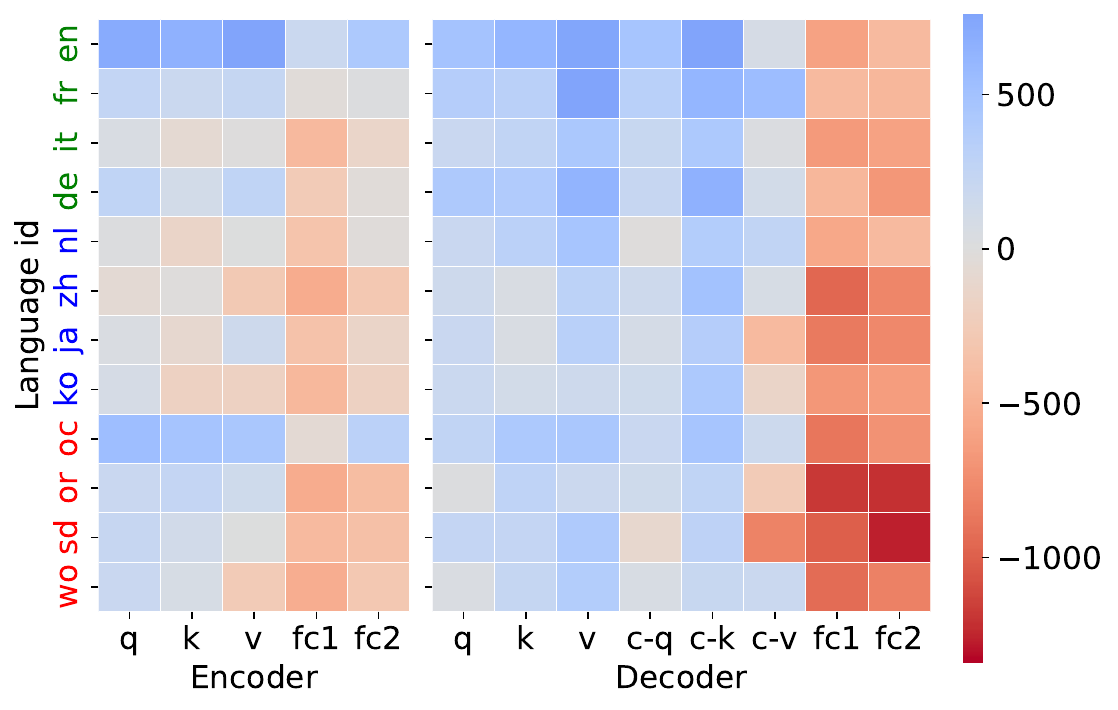}
        \caption{Layer 9}
    \end{subfigure}
    \begin{subfigure}{0.66\columnwidth}
        \includegraphics[width=\columnwidth]{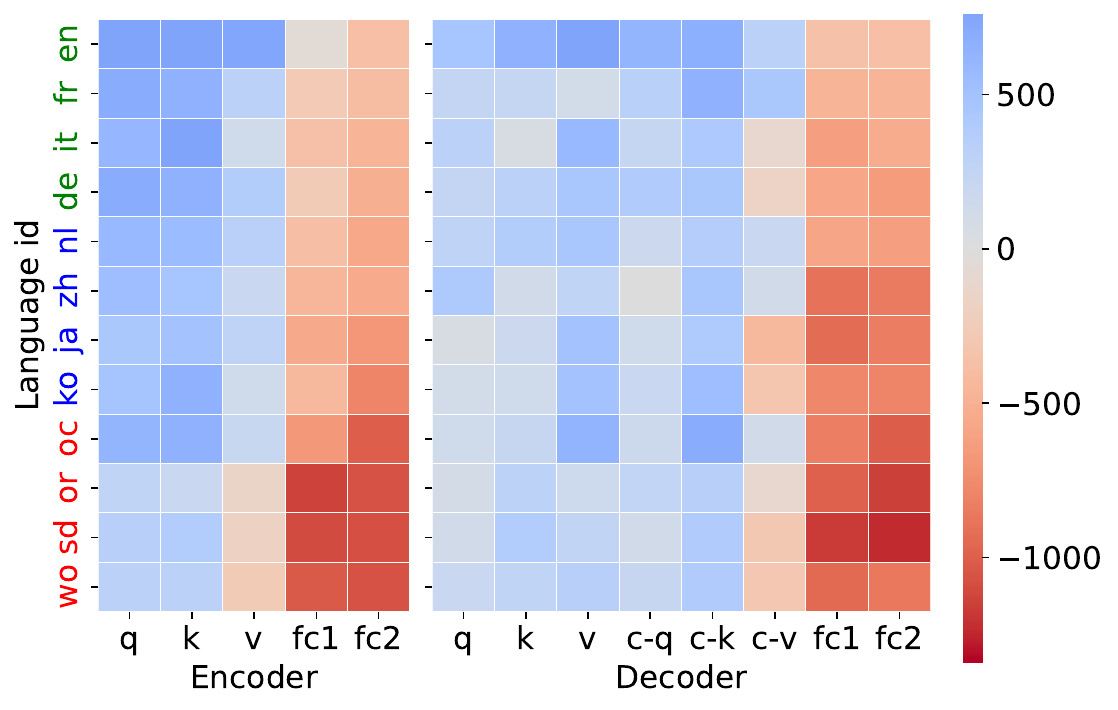}
        \caption{Layer 10}
    \end{subfigure}
    \hfill
    \begin{subfigure}{0.66\columnwidth}
        \includegraphics[width=\columnwidth]{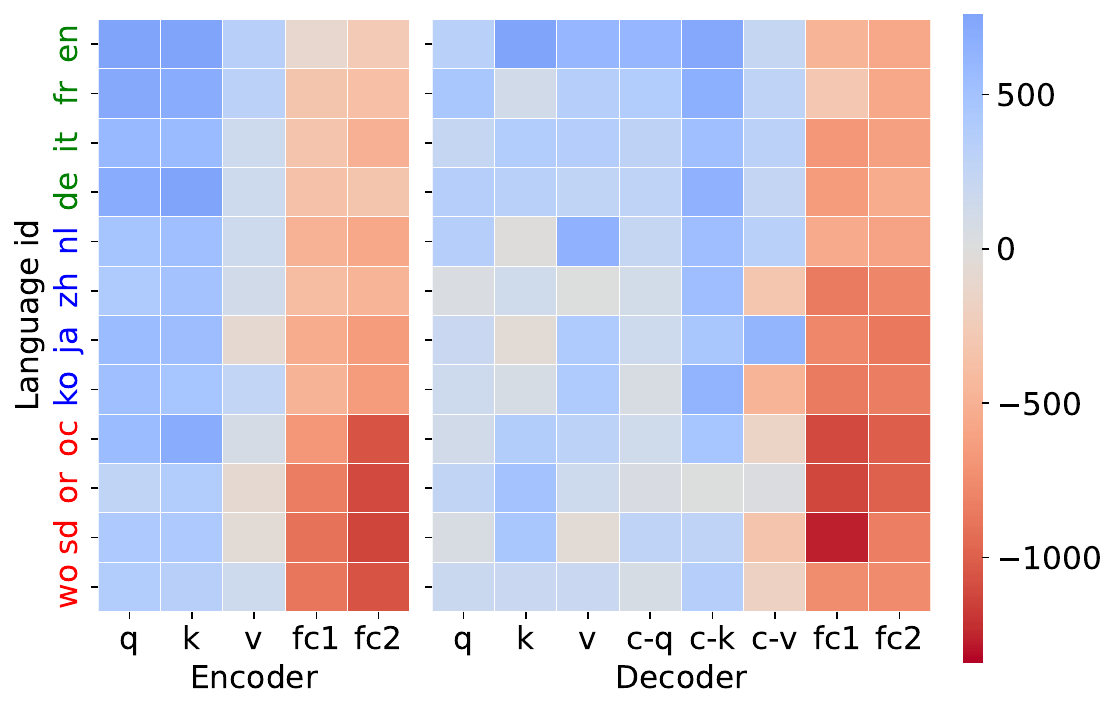}
        \caption{Layer 11}
    \end{subfigure}
    \hfill
    \begin{subfigure}{0.66\columnwidth}
        \includegraphics[width=\columnwidth]{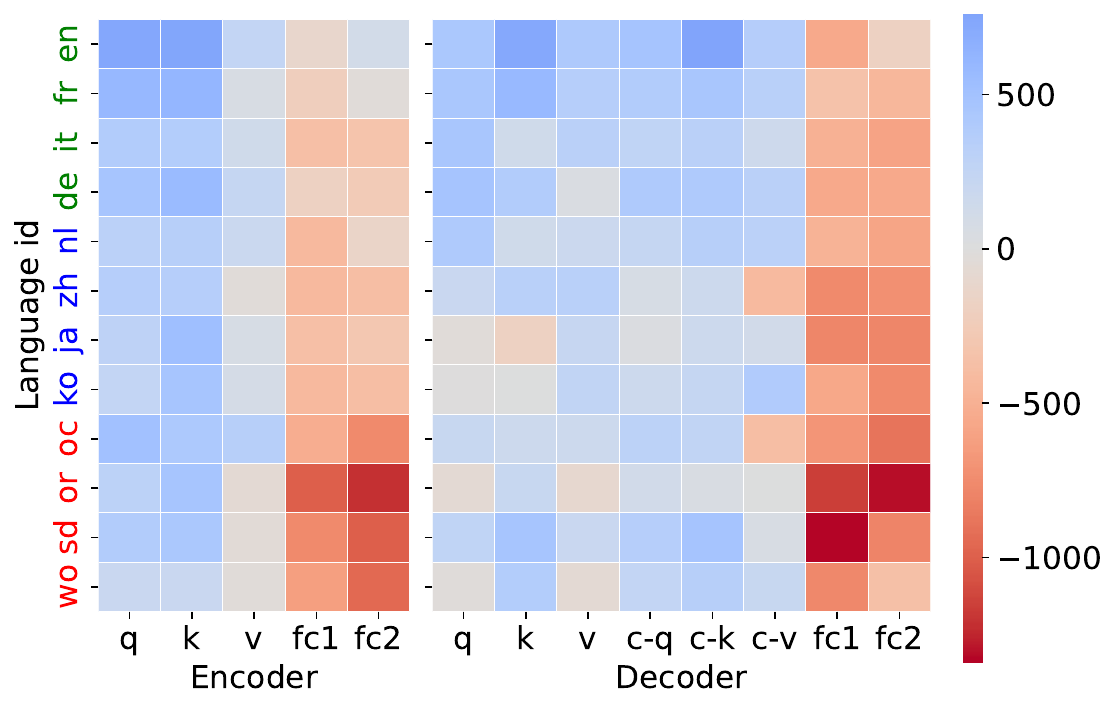}
        \caption{Layer 12}
    \end{subfigure}
    \caption{The parameter space demands for each language in all 12 layers of encoder and decoder respectively. Red color means a higher demand. We can see a clear tendency that very-low-resource languages require more parameters during fine-tuning.}
    \label{fig:ise-pruning}
\end{figure*}

We present the results of Intrinsic Subspace Estimation in all 12 layers of encoder and decoder in Figure \ref{fig:ise-pruning}. The results show a clear tendency that the required subspace size for each language is highly correlated with its resource type. Very-low-resource languages require more parameters for fine-tuning compared to high and medium-resource languages. To more concretely measure the observed relationship, we calculate the correlation \footnote{\url{https://docs.scipy.org/doc/scipy/reference/generated/scipy.stats.pearsonr.html}} between the number of available Bitext with English in Table \ref{tab:lang12} and the importance score for each module. As shown in following Table \ref{tab:ccEncoder} and Table \ref{tab:ccDecoder}, the results show a positive correlation between the resource level and our proposed importance score (the lower the resource, the lower the importance score, and the higher the parameter space demands).

\section{Language-specific Pruning}
\label{app:lspruning}
\begin{figure*}[t]
    \begin{subfigure}{0.66\columnwidth}
        \includegraphics[width=\columnwidth]{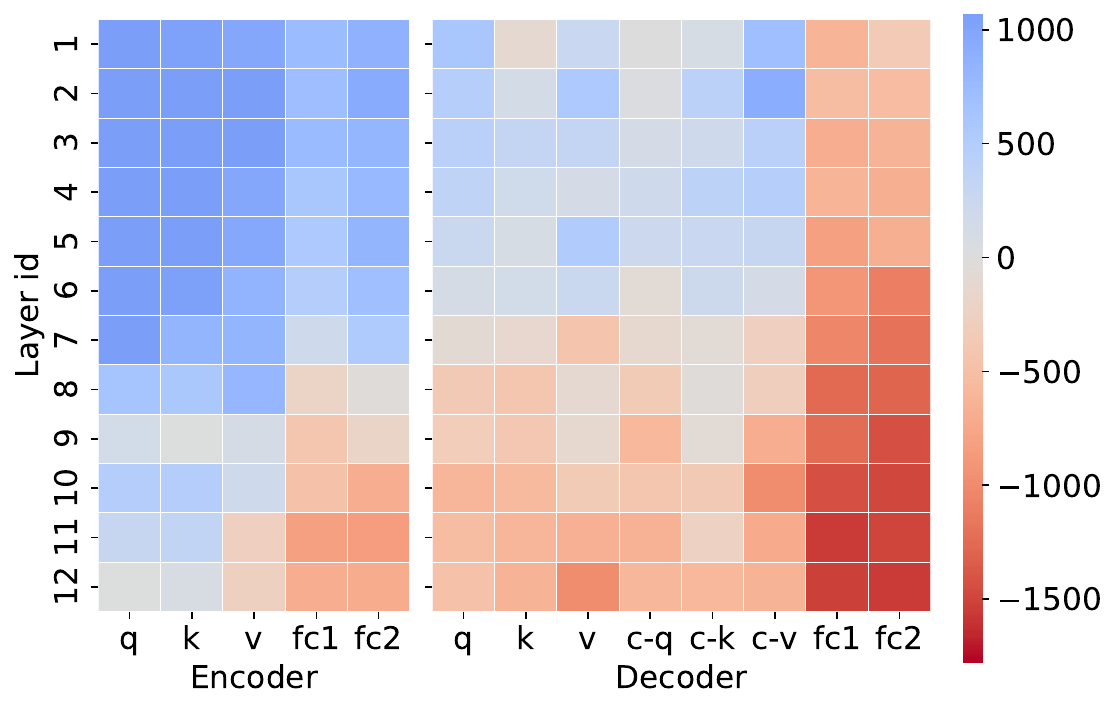}
        \caption{de}
    \end{subfigure}
    \begin{subfigure}{0.66\columnwidth}
        \includegraphics[width=\columnwidth]{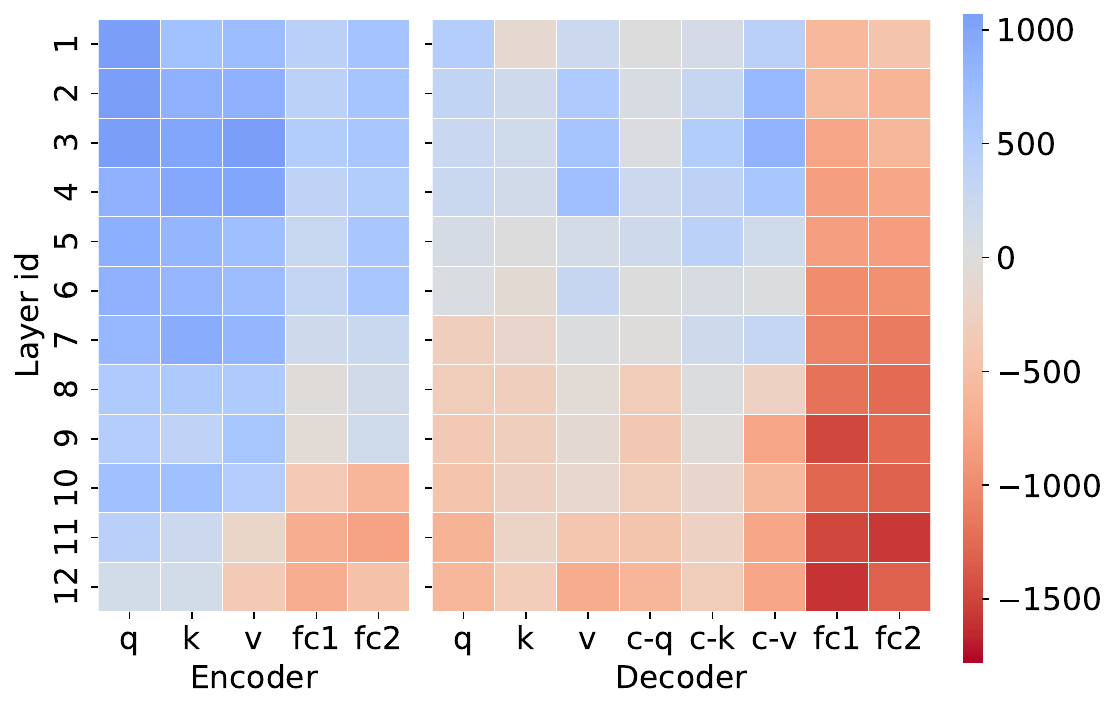}
        \caption{en}
    \end{subfigure}
    \begin{subfigure}{0.66\columnwidth}
        \includegraphics[width=\columnwidth]{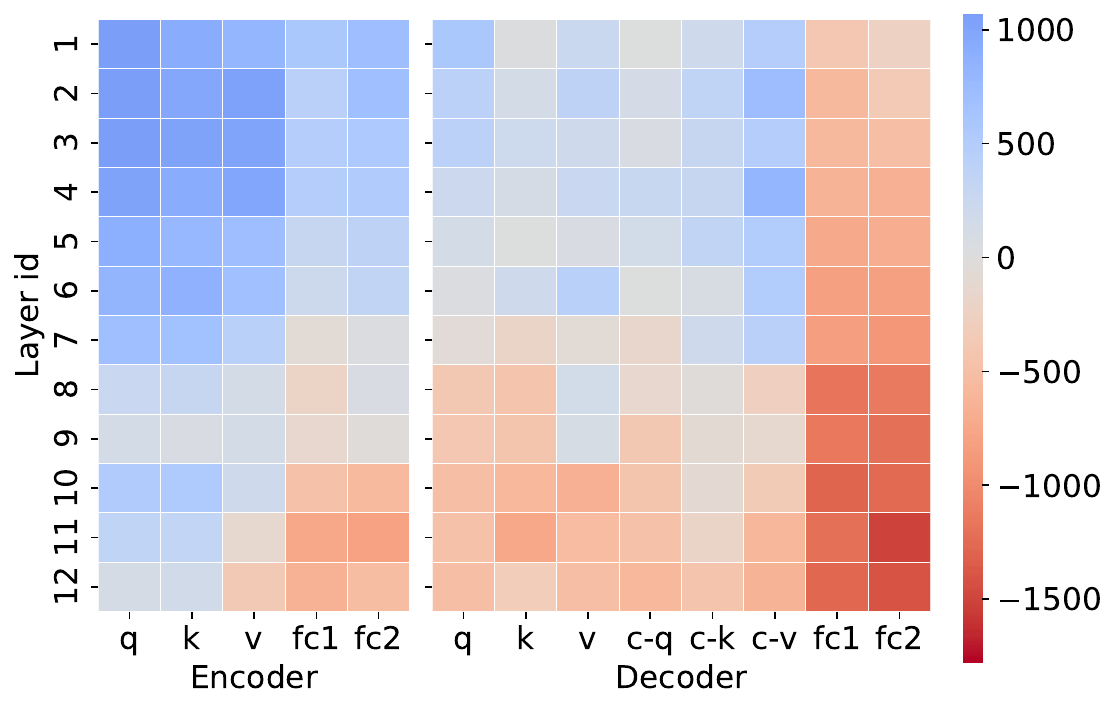}
        \caption{fr}
    \end{subfigure}
    \begin{subfigure}{0.66\columnwidth}
        \includegraphics[width=\columnwidth]{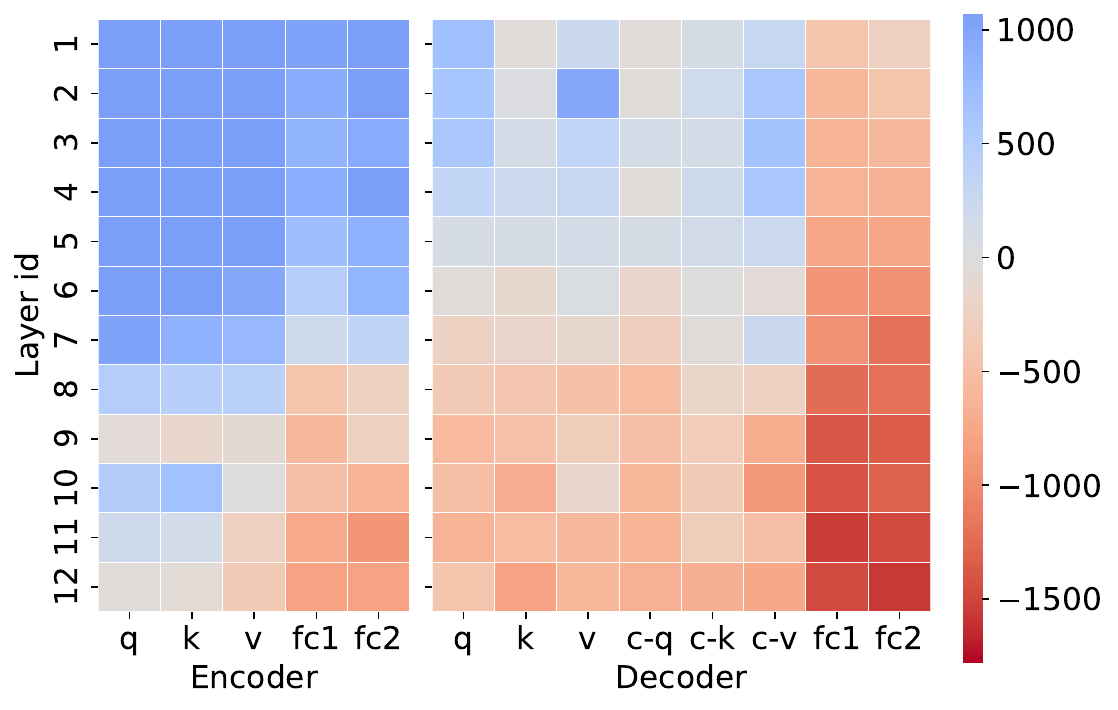}
        \caption{it}
    \end{subfigure}
    \begin{subfigure}{0.66\columnwidth}
        \includegraphics[width=\columnwidth]{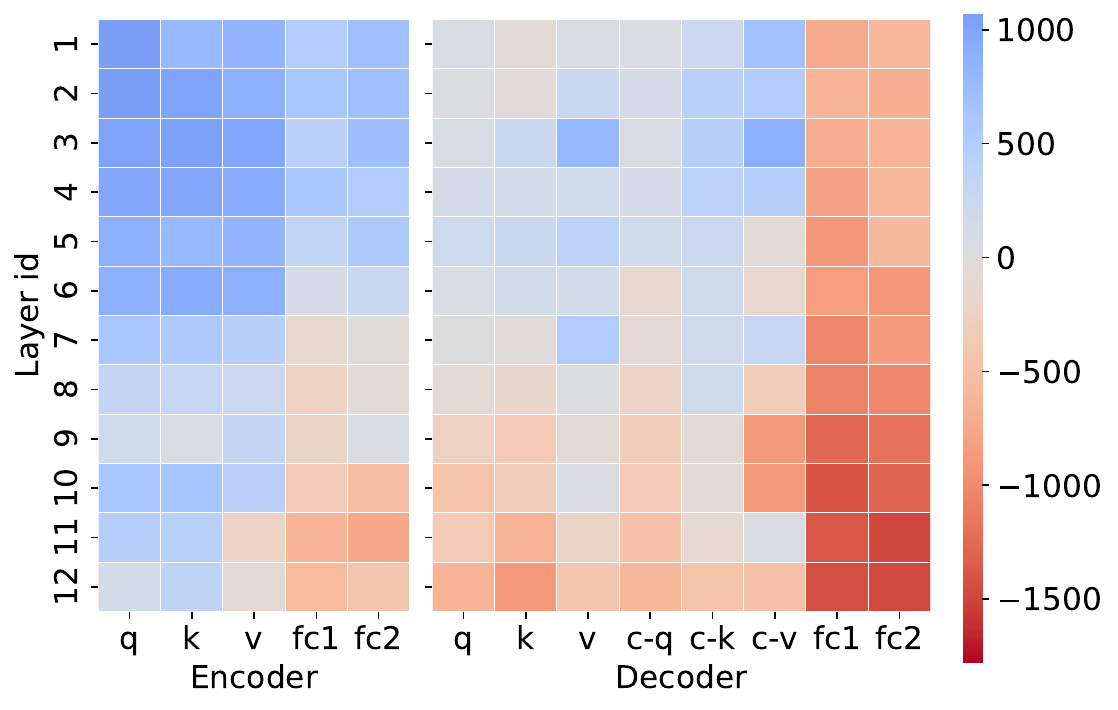}
        \caption{ja}
    \end{subfigure}
    \begin{subfigure}{0.66\columnwidth}
        \includegraphics[width=\columnwidth]{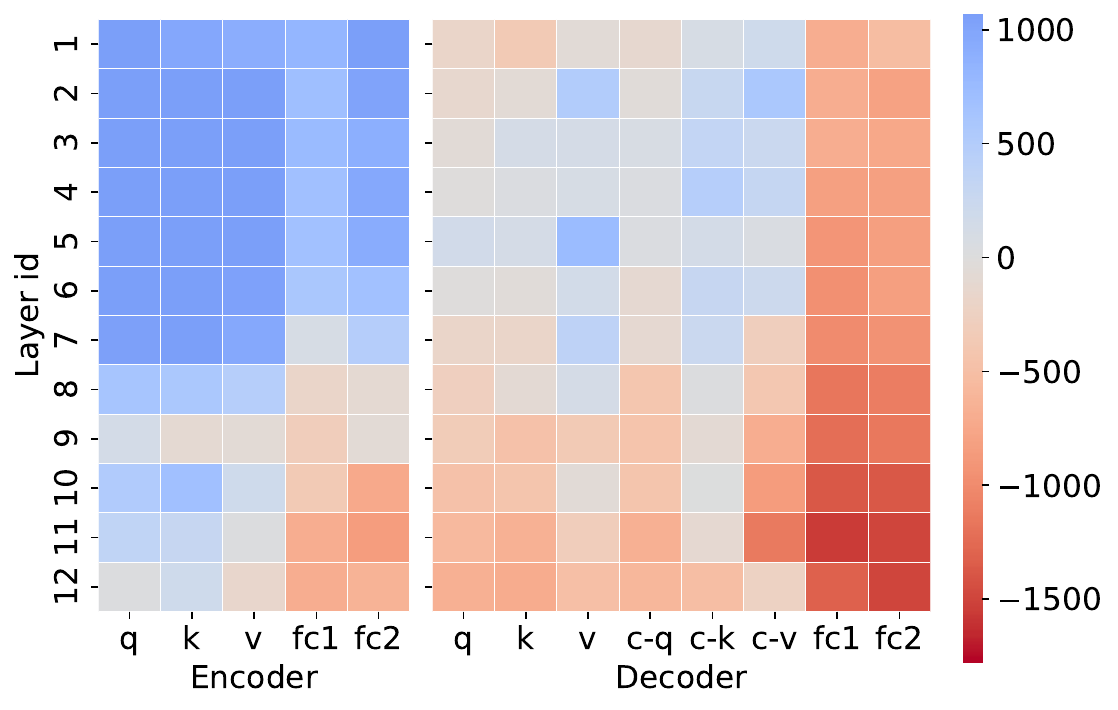}
        \caption{ko}
    \end{subfigure}
    \hfill
    \begin{subfigure}{0.66\columnwidth}
        \includegraphics[width=\columnwidth]{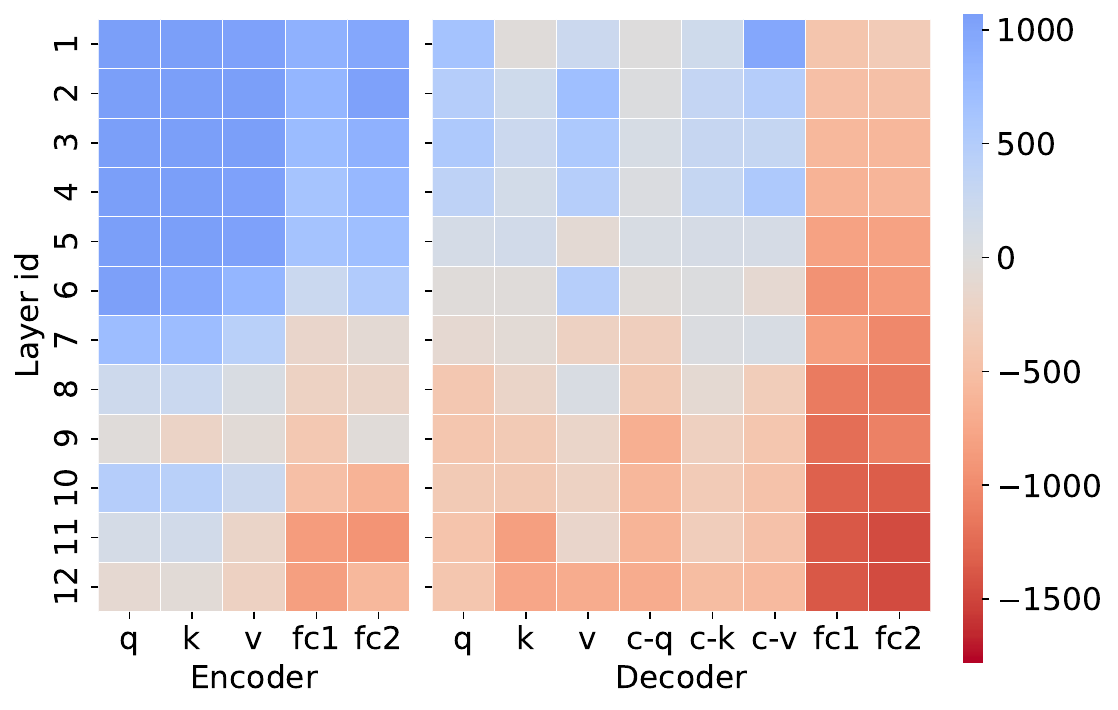}
        \caption{nl}
    \end{subfigure}
    \begin{subfigure}{0.66\columnwidth}
        \includegraphics[width=\columnwidth]{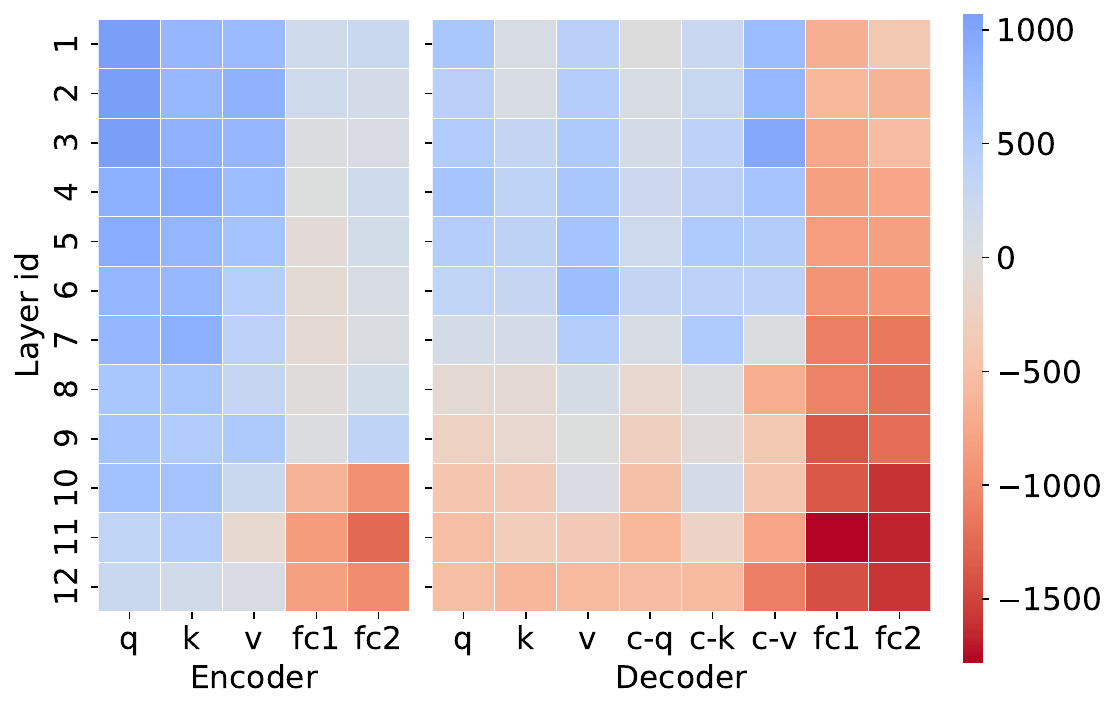}
        \caption{oc}
    \end{subfigure}
    \begin{subfigure}{0.66\columnwidth}
        \includegraphics[width=\columnwidth]{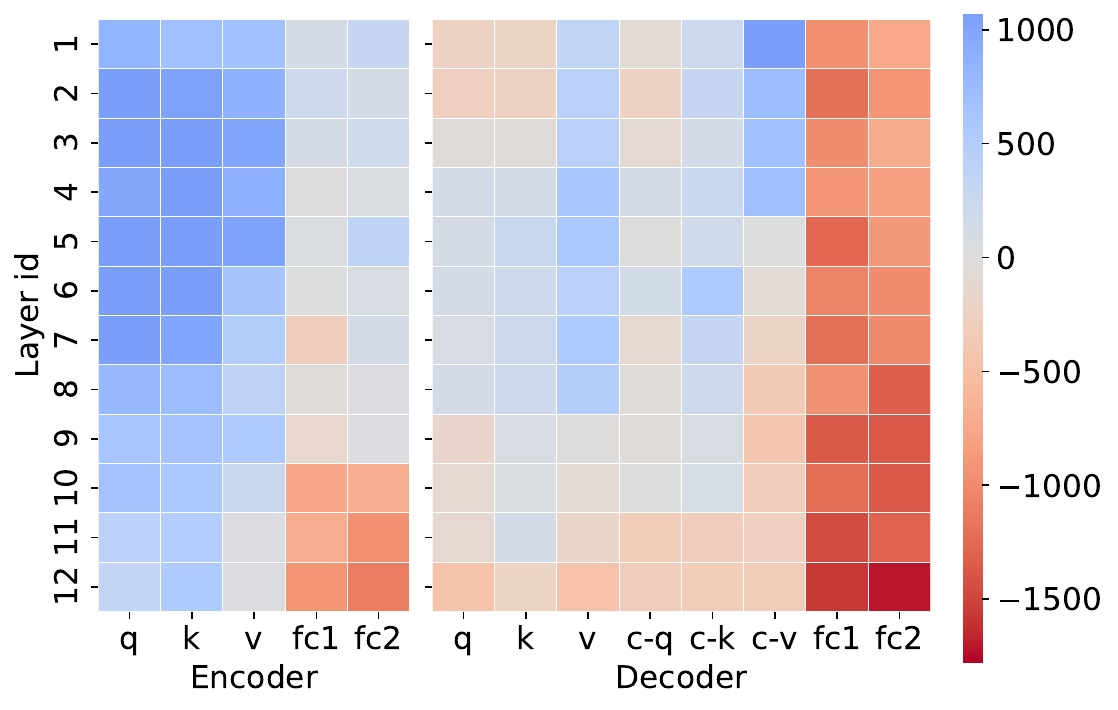}
        \caption{or}
    \end{subfigure}
    \begin{subfigure}{0.66\columnwidth}
        \includegraphics[width=\columnwidth]{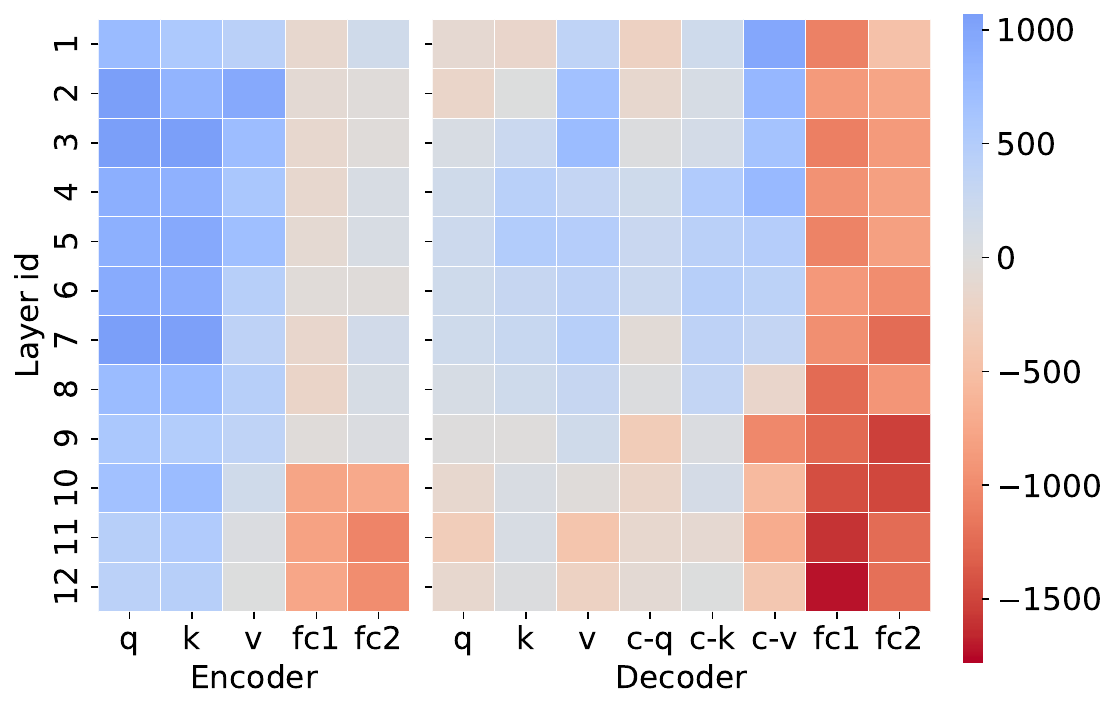}
        \caption{sd}
    \end{subfigure}
    \hfill
    \begin{subfigure}{0.66\columnwidth}
        \includegraphics[width=\columnwidth]{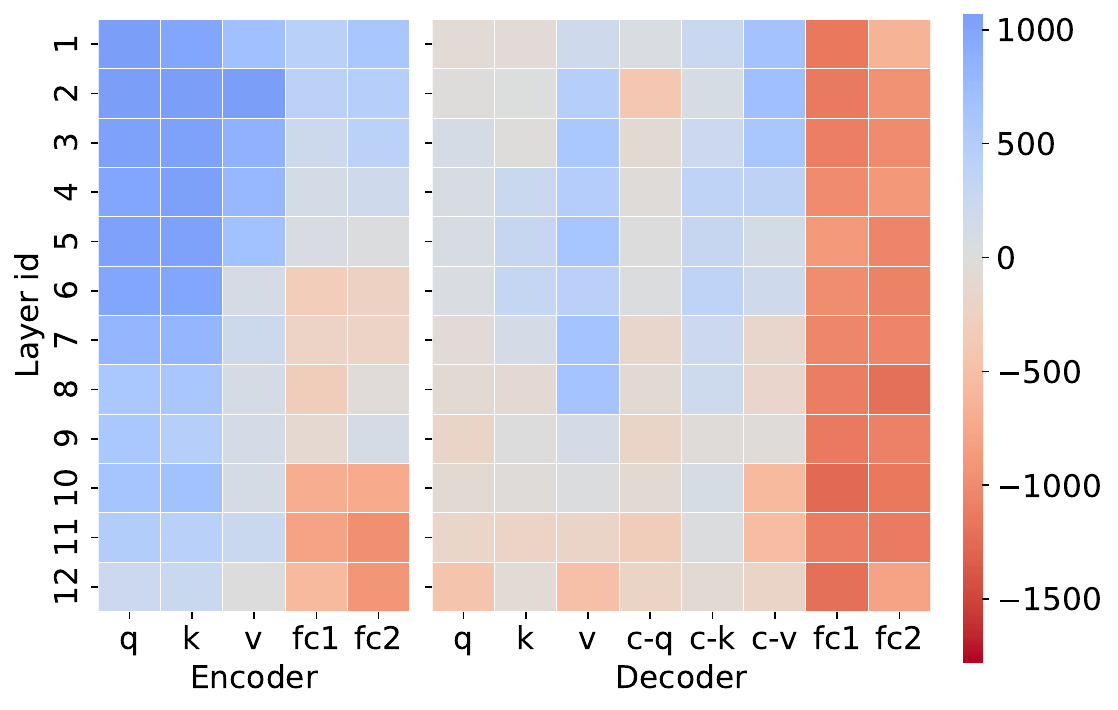}
        \caption{wo}
    \end{subfigure}
    \hfill
    \begin{subfigure}{0.66\columnwidth}
        \includegraphics[width=\columnwidth]{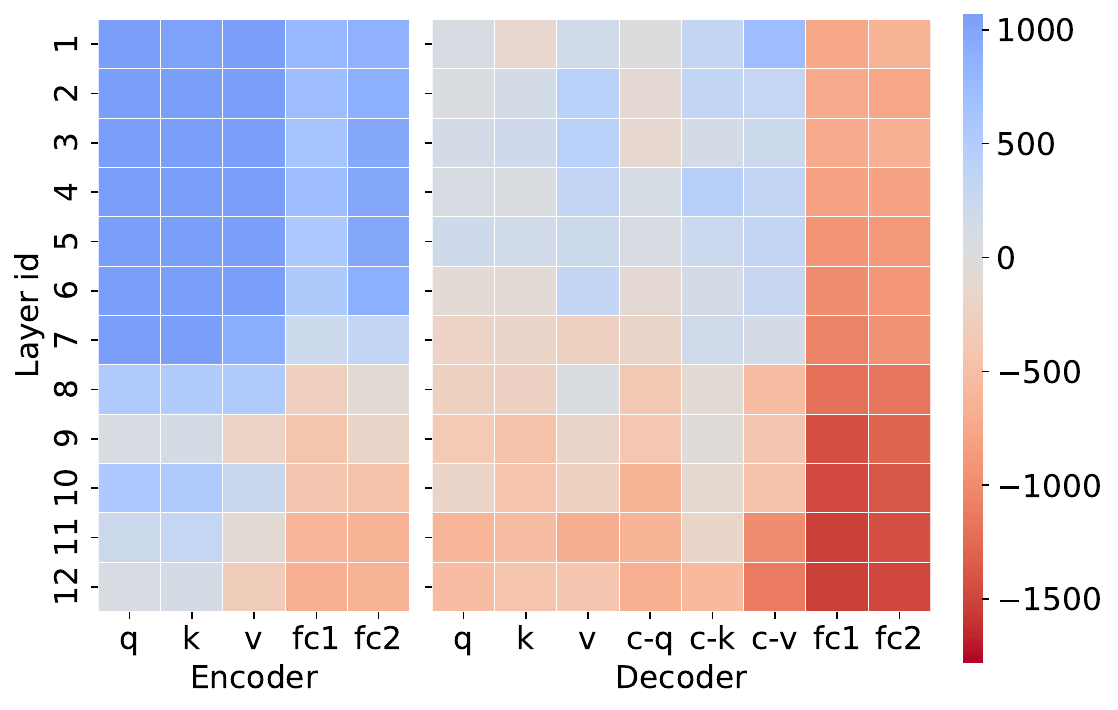}
        \caption{zh}
    \end{subfigure}
    \caption{Parameter space demands of different languages in encoder and decoder respectively. Red color means a higher
demand. We can see a clear trend across all languages that fc1 and fc2 in the top layers of the encoder are more important than other weight matrices.}
    \label{fig:lang-pruning}
\end{figure*}

Language-specific pruning is applied to analyze the importance of different weight matrices for each language. We add LSLo with a rank of 8 to all weight matrices. Given $n$ languages, each LSLo module will have $n$ language-specific LoRA modules. All $B$ matrices of LoRA are divided into $n$ groups by language. By applying global pruning to each group, we can analyze which weight matrices are most important for each language. As shown in Figure \ref{fig:lang-pruning}, we can see a clear tendency among all languages that fc1 and fc2 play a more important role than other weight matrices.

\end{document}